\documentclass[journal]{IEEEtran}
\renewcommand{\baselinestretch}{1.0}
\renewcommand{\theenumii}{\theenumi.\arabic{enumii}}

\usepackage[utf8]{inputenc}
\usepackage{subfigure}
\usepackage{amssymb}
\usepackage{color,soul}
\usepackage[T1]{fontenc}
\usepackage{multicol}
\usepackage{multirow}
\usepackage{textcomp}
\usepackage{multirow}
\usepackage{bigdelim}
\usepackage{graphicx}
\usepackage{epstopdf}
\usepackage{amssymb}
\usepackage{amsmath}
\usepackage{psfrag}
\usepackage{amsthm}
\usepackage{subfigure}
\usepackage{algorithmic,algorithm}
\usepackage{mathrsfs}
\usepackage{framed}
\usepackage{color}
\usepackage{multirow,enumerate}
\usepackage{bigdelim}
\definecolor{shadecolor}{rgb}{0.92,0.92,0.92}
\usepackage{soul}
\usepackage{color}
\usepackage{cuted}
\usepackage{cite}
\theoremstyle{definition}

\newtheorem{definition}{Definition}

\newtheorem{remark}{Remark}

\makeatletter
\newcommand{\vast}{\bBigg@{3.2}}
\newcommand{\Vast}{\bBigg@{4.5}}

\makeatother

\graphicspath{{Figure/}}

\usepackage{etoolbox}
\makeatletter 
\pretocmd\@bibitem{\color{black}\csname keycolor#1\endcsname}{}{\fail}
\newcommand\citecolor[1]{\@namedef{keycolor#1}{\color{blue}}}
\makeatother

\begin{document}
	
	\title{Alternating Direction Method of Multipliers for Constrained Iterative LQR in Autonomous Driving}
	
	\author{
		Jun Ma, 
		Zilong Cheng, 
		Xiaoxue Zhang, 
		Masayoshi Tomizuka,  \IEEEmembership{Life Fellow,~IEEE,}
		and Tong Heng Lee	
\thanks{J. Ma is with the Robotics and Autonomous Systems Thrust, The Hong Kong University of Science and Technology (Guangzhou), Guangzhou, China, with the Department of Electronic and Computer Engineering, The Hong Kong University of Science and Technology, Hong Kong SAR, China, and also with the HKUST Shenzhen-Hong Kong Collaborative Innovation Research Institute, Futian, Shenzhen, China (e-mail: jun.ma@ust.hk).}
		\thanks{Z. Cheng, X. Zhang, and T. H. Lee are with the NUS Graduate School for Integrative Sciences and Engineering, National University of Singapore, 119077 (e-mail: zilongcheng@u.nus.edu; xiaoxuezhang@u.nus.edu; eleleeth@nus.edu.sg).}
		\thanks{M. Tomizuka is with the Department of Mechanical Engineering, University of California, Berkeley, CA 94720 USA (e-mail: tomizuka@berkeley.edu).}
		\thanks{This work has been submitted to the IEEE for possible publication. Copyright may be transferred without notice,	after which this version may no longer be accessible.}
	}
	
	\maketitle

	\begin{abstract}
		In the context of autonomous driving, the iterative linear quadratic regulator (iLQR) is known to be an efficient approach to deal with the nonlinear vehicle model in motion planning problems. Particularly, the constrained iLQR algorithm has shown noteworthy advantageous outcomes of computation efficiency in achieving motion planning tasks under general constraints of different types. However, the constrained iLQR methodology requires a feasible trajectory at the first iteration as a prerequisite when the logarithmic barrier function is used. Also, the methodology leaves open the possibility for incorporation of fast, efficient, and effective optimization methods (i.e., fast-solvers) to further speed up the optimization process such that the requirements of real-time implementation can be successfully fulfilled. In this paper, a well-defined and commonly-encountered motion planning problem is formulated under nonlinear vehicle dynamics and various constraints, and an alternating direction method of multipliers (ADMM) is utilized to determine the optimal control actions leveraging the iLQR. With this development, the approach is able to circumvent the feasibility requirement of the trajectory at the first iteration. An illustrative example of motion planning for autonomous vehicles is then investigated with different driving scenarios taken into consideration. 
A noteworthy achievement of high computation efficiency is attained with the proposed development; comparing with the constrained iLQR algorithm based on the logarithmic barrier function, our proposed method reduces the average computation time by $31.93\%$, $38.52\%$, and $44.57\%$ in the three scenarios; compared with the optimization solver IPOPT, our proposed method reduces the average computation time by $46.02\%$, $53.26\%$, and $88.43\%$ in the three scenarios. As a result, real-time computation and implementation can be realized through our proposed framework, and thus it provides additional safety to the on-road driving tasks. 
	\end{abstract}
	 
	
	\section{Introduction}
	Due to the rapid increase in traffic density, vehicle safety has become a primary concern in modern intelligent transportation systems. As one of the promising approaches to relieve traffic problems and enhance driving safety, autonomous vehicles have demonstrated great potentials in the current vehicle technology~\cite{li2017dynamical,zhang2011data}. Consequently, substantial efforts have been devoted to research on autonomous driving in recent years~\cite{hang2021cooperative,hang2022driving,huang2021human,chen2021exploring,li2022stepwise}. Particularly, as one of the core areas in autonomous driving, motion planning has attracted the attention of several disciplines, where it aims to plan a feasible trajectory that conforms to the requirements in terms of obstacle avoidance, energy consumption, traveling time, etc. In fact, there are a number of traditionally challenging dynamic motion planning problems in transportation, which have long threads of research literature~\cite{wang2010parallel,gonzalez2015review}. However, they have readily stood to benefit tremendously from the recent advances in optimization and artificial intelligence~\cite{hou2019interactive,duan2021distributional,chen2021interpretable}. 
	
	Along with this line, the planning methods in the continuous space generally include the learning-based method and the optimization-based method. For the learning-based method, autonomous vehicles can continuously improve their proficiency from the outcomes of navigational decisions~\cite{duan2020hierarchical,chen2020reinforcement}. Nonetheless, the lack of a theoretical guarantee causes potential safety concerns. On the other hand, the optimization-based method relies on the formulation of a mathematical optimization problem, where system requirements can be explicitly expressed as equality and inequality constraints as part of the model predictive control (MPC) synthesis. To be more specific, typical constraints involved in the optimization problem include the dynamic constraint, environmental constraint (such as the obstacle avoidance constraint), physical limit constraint (such as the constraint on the steering angle, acceleration of the engine, and deceleration of the brake), etc. For the optimization-based method, the MPC has been widely deployed to solve the sequential decision-making problems with certain cumulative objectives considered over a certain horizon~\cite{mayne2000constrained,liu2017path}. In view of its appropriate applicability, various research works present the use of MPC-based methods to generate a feasible solution in the trajectory generation problem~\cite{ji2016path}. However, most of these MPC-based works focus on simple motion tasks with simplified vehicle models. In more realistic and complex situations, the MPC is no longer effective due to the nonlinearity of the vehicle model. Moreover, the generated trajectories need to deal with highly dynamic and complex driving scenarios (such as lane change, maneuvering, turning, overtaking, etc.)  in a spatiotemporal domain. As a result, the high complexity and non-convexity of these constraints incurred add additional burdens to the computation efficiency of the MPC~\cite{zhang2019integrated,zhang2020trajectory}.
	
	Due to the limitation of the MPC in handling the nonlinear systems, the differential dynamic programming (DDP) is developed, which is known as a second-order shooting method admitting quadratic convergence~\cite{mayne1966second,xie2017differential}. Classical DDP needs the second-order derivatives of the dynamics, yet the determination of the second-order derivatives aggravates the computation burden. Under a special condition that only the first-order derivative is used, it is reduced to the so-called the iterative linear quadratic regulator (iLQR) method with the Gauss-Newton approximation~\cite{nagariya2020iterative}. Basically, the iLQR expands the idea of the conventional LQR from linear systems to nonlinear systems, and it is known as an optimization-based method for nonlinear systems. Compared with the LQR, the iLQR optimizes a whole control sequence instead of just the control action at the current time. In the iLQR architecture, an initial trajectory is defined and then the iLQR is executed to refine the trajectory in an iterative framework, such that the optimal solution can be obtained efficiently. It is pertinent to note that the conventional iLQR only takes the system dynamic constraint into account in the optimization process, and it is the major shortcoming of the iLQR that it cannot handle the general inequality constraints appropriately. 
	Therefore, to overcome this rather significant impediment, control-limited differential dynamic programming has been used to cater to the box constraints imposed on the control input~\cite{tassa2014control}. In addition, an innovative variant of the iLQR so-called the constrained iLQR has been developed, which offers the inclusion of various general constraints~\cite{pan2020safe,chen2019autonomous}. Rather importantly, the constrained iLQR demonstrates noteworthy performance in terms of performance and computation efficiency, despite the existence of very involved nonlinear characteristics of the vehicle model and non-convexity of the obstacle avoidance constraint. Notably, the barrier function is adopted in~\cite{chen2019autonomous} such that the general constraints are incorporated into the objective function appropriately. It is remarkable that it has shown significant improvement in computation efficiency as compared with the sequential quadratic programming (SQP) solver. However, the use of the logarithmic barrier function and the outer-inner loop framework invokes additional iterations. Also, it requires a feasible trajectory at the first iteration by using the iLQR with the logarithmic barrier function. Moreover, the optimization solver named IPOPT is widely deployed to solve such optimization problems. With this method, the optimization problem is solved directly with all predefined constraints coped simultaneously, and it usually takes longer time to find a satisfying solution. Hence, it remains an open problem for further improvement of the computation efficiency to meet the need for real-time implementation on top of that; and to this extent, the autonomous vehicle can better respond to emergencies in practical situations.
	
	Nowadays, the development of numerical optimization tools has progressed leaps and bounds. As an emerging technique with excellent scalability, the alternating direction method of multipliers (ADMM)~\cite{boyd2011distributed} has been deployed successfully in various domains, including optimal control, distributed computation, machine learning, and so on~\cite{ma2020symmetric,raja2020sp,cheng2020semi,ma2021optimal}. Essentially, the ADMM extends the method of multipliers and splits the optimization variable into two parts, and then the invoked sub-problems resulting from the splitting schemes can be attempted and solved in a separate framework. This approach relieves the typical computation burden arising from the growth of system dimensions. Recent advances in the ADMM enable an effective determination of the global optimum of a convex optimization problem, and these advances also motivate the researchers to investigate the behavior of the ADMM in non-convex problems. For example, \cite{xu2016empirical} gives an empirical study of the ADMM for non-convex problems, and \cite{hong2016convergence} performs the convergence analysis of the ADMM for a family of non-convex problems. Furthermore, the convergence behavior of the ADMM for problems with nonlinear equality constraints is investigated in \cite{wang2017nonconvex}. Also, the convergence conditions for a coupled objective function with non-convex and non-smooth characteristics are studied in \cite{wang2019global}. In these past research works, the ADMM has been reasonably established at the theoretical level. Certainly, these advanced optimization techniques bring promising prospects to the area of autonomous driving.
	
	This paper presents an innovative and effective ADMM-based constrained iLQR approach for motion planning. In this work, the nonlinearity of the vehicle model is considered in the problem formulation. Also, various constraints are suitably addressed in this work, including the  constraint on the acceleration due to the engine force limit and the braking force limit, the constraint on the steering angle due to the mechanical limitation of the vehicle, and the obstacle avoidance constraint for safety concern in driving scenarios. The main contributions of this work are:
	
	\begin{itemize}
		\item {A novel constrained iLQR approach is presented to deal with the aforementioned constraints incurred in autonomous driving applications, and the nonlinearity in the vehicle system dynamics is addressed effectively.}
		
		\item 
			The ADMM algorithm is utilized to split the optimization problem into several manageable sub-problems. 
			Therefore, the computation burden resulting from the nonlinearity and non-convexity of the motion planning problem is alleviated. The efficient computation makes the real-time implementation possible, and thus extra assurance of driving safety is provided.
		
		\item {The proposed algorithm also avoids the necessity of searching for a feasible trajectory at the first iteration, which allows for more flexibility to set the initial trajectory.}
	\end{itemize}

	The remainder of this paper is organized as follows. Section II presents the vehicle model, objective function, and constraints in the autonomous driving task. Section III presents the ADMM-based constrained iLQR approach for motion planning. In Section IV, an illustrative example in autonomous driving is given to show the effectiveness of the proposed methodology. At last, the conclusions of this work are given in Section V.

	\section{Problem Statement}
	\subsection{Vehicle Model}

\begin{figure}[t]
				\centering
				\includegraphics[trim=0 0 0 0, clip, width=0.85\linewidth]{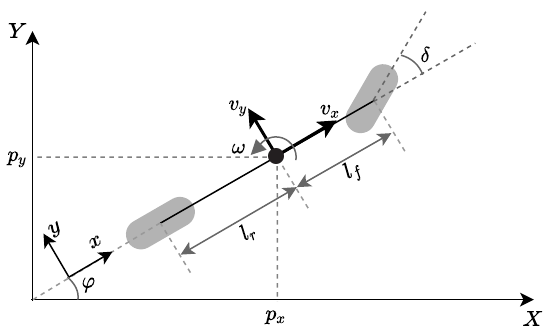}
				\caption{Illustration of the vehicle model.}
				\label{fig:vehimodel}
			\end{figure}

		As shown in Fig.~\ref{fig:vehimodel}, the bicycle model used in~\cite{ge2020numerically} is borrowed to represent the vehicle dynamics. In this model, the aerodynamic influences, slip phenomena, and suspension movements are neglected.
		In terms of the vehicle model used in this work, the state vector is defined as $x=\begin{bmatrix}
					p_x & p_y & \varphi &v_x  & v_y
					& \omega	\end{bmatrix}^{\top}$, where $p_x$ and $p_y$ represent the X-coordinate and Y-coordinate of the center point of the vehicle, respectively; $\varphi$ denotes the heading angle; $v_x$ and $v_y$ denote the longitudinal and lateral velocity of the vehicle; $\omega$ is the yaw rate.  Also, the control input vector is defined as $u=\begin{bmatrix}
					a & \delta
				\end{bmatrix}^{\top}$, where $a$ and $\delta$ denote the acceleration and the steering angle, respectively. Here, we also define the mass of the vehicle as $m$, $l_f$ and $l_r$ as the distance from the center of mass to the front and rear axle, $k_f$ and $k_r$ as the cornering stiffness of the front and rear wheels, and $I_{z}$ as the polar moment of inertia. For simplicity, we further define $L_k=l_{f} k_{f}-l_{r} k_{r}$. 
			
		Given the time step $T_s$, the nonlinear discrete model of the vehicle is expressed as
				\begin{IEEEeqnarray}{rCl}
					&&\quad\left[\begin{array}{c}
						p_x(\tau+1) \\
						p_y(\tau+1) \\
						\varphi_(\tau+1) \\
						v_x(\tau+1) \\
						v_y(\tau+1) \\
						\omega(\tau+1)
					\end{array}\right]\triangleq f\big(x(\tau),u(\tau)\big)\nonumber\\&&=\left[\begin{array}{c}
						p_x(\tau)+T_{s}\left(v_x(\tau) \cos \varphi(\tau)-v_y(\tau) \sin \varphi(\tau)\right) \\
						p_y(\tau)+T_{s}\left(v_y(\tau) \cos \varphi(\tau)+v_x(\tau) \sin \varphi(\tau)\right) \\
						\varphi(\tau)+T_{s} \omega(\tau) \\
						v_x(\tau)+T_{s} a(\tau) \\
						\frac{m v_x(\tau) v_y(\tau)+T_{s}L_k \omega(\tau)-T_{s} k_{f} \delta(\tau) v_x(\tau)-T_{s} m v_x(\tau)^{2} \omega(\tau)}{m v_x(\tau)-T_{s}\left(k_{f}+k_{r}\right)} \\
						\frac{I_{z} v_x(\tau) \omega(\tau)+T_{s}L_k v_y(\tau)-T_{s} l_{f} k_{f} \delta(\tau) v_x(\tau)}{I_{z} v_x(\tau)-T_{s}\left(l_{f}^{2} k_{f}+l_{r}^{2} k_{r}\right)}
					\end{array}\right]. \nonumber\\ \IEEEeqnarraynumspace
				\end{IEEEeqnarray}

			\subsection{Objective Function}
			In the objective function, the position tracking error and longitudinal velocity tracking error are considered separately. Also, the steering wheel input is considered because it is closely related to the safety and comfort of passengers in on-road driving scenarios. Moreover, the acceleration is taken into account, as it reflects the fuel consumption as well as the comfort of passengers. With the above descriptions, the following objective function in each time stamp is constructed:
				\begin{IEEEeqnarray}{rCl}
					&&\ell_\tau\big(x(\tau),u(\tau)\big)\nonumber\\ &=&    d_{q_1,q_2}\big(x(\tau),l^r(\tau)\big)+ q_3 \Big\|M_{v_x} x(\tau)-v^r_{x}(\tau)\Big\|^2\nonumber \\
					&&+ r_1 u(\tau)^{\top} M_\delta u(\tau) +r_2  u(\tau)^{\top} M_a u(\tau),\IEEEeqnarraynumspace
				\end{IEEEeqnarray}	
			where $\|\cdot\|$ denotes the Euclidean norm operator; $q_1$, $q_2$, $q_3$, $r_1$, and $r_2$ represent the weighting parameters of the X-coordinate position tracking error, Y-coordinate position tracking error, longitudinal velocity tracking error, steering wheel input, and acceleration, respectively; $l^r$ denotes the polyline of the reference trajectory; $d_{q_1,q_2}$ denotes the weighted Euclidean distance between the vehicle and the corresponding polyline; $v^r_{x}$ denotes the reference of longitudinal velocity; $M_{v_x}$ is the matrix to extract the longitudinal velocity of the vehicle from the state vector, i.e., $M_{v_x}=\begin{bmatrix}
				0&0&0&1&0&0
			\end{bmatrix}$; $M_\delta$ and $M_a$ are two matrices given by $M_\delta = \begin{bmatrix}
				0& 0 \\0 & 1
			\end{bmatrix}$ and $M_a = \begin{bmatrix}
				1& 0 \\0 & 0
			\end{bmatrix}$.
			
			In this work, to facilitate the use of the iLQR method, the objective function at each time stamp $\tau$ is transformed into the following form:
				\begin{IEEEeqnarray}{rCl}
					\ell_\tau\big(x(\tau),u(\tau)\big) = \begin{bmatrix}
						x(\tau) \\ u(\tau)
					\end{bmatrix}^{\top} C \begin{bmatrix}
						x(\tau) \\ u(\tau)
					\end{bmatrix} -2 \begin{bmatrix}
						x(\tau) \\ u(\tau)
					\end{bmatrix}^{\top} C r+\textup{const},\nonumber\\ 
				\end{IEEEeqnarray}
				where \begin{IEEEeqnarray}{rCl}
					C&=&\text{diag}\{q_1,q_2,0,q_3,0,0,r_1,r_2\},\nonumber\\ 
					r&=&\begin{bmatrix}
						p_{x}^r & p_{y}^r & 0& v_{x}^r& 0 & 0 & 0 & 0
					\end{bmatrix}^{\top},
				\end{IEEEeqnarray} 
				where $p_{x}^r$, $p_{y}^r$, $v_{x}^r$ denote the X-coordinate reference position, Y-coordinate reference position, and reference longitudinal velocity, respectively.  
			
			\subsection{Constraints}
			First of all, the steering angle of the vehicle is bounded due to its mechanical limitations. Therefore, for all $\tau=0,1,\dotsm,T-1$, we have
			\begin{IEEEeqnarray}{rCl}
				-\delta_{max} \leq V_\delta u(\tau) \leq \delta_{max},
			\end{IEEEeqnarray}
			with $ V_\delta = \begin{bmatrix}
				0&1 
			\end{bmatrix}$, and $\delta_{max}$ represents the largest possible steering angle that the vehicle attains. 
			
			Then, due to the engine force limit and braking force limit of the vehicle, the acceleration is also bounded, that is for all $\tau=0,1,\dotsm,T-1$,
			\begin{IEEEeqnarray}{rCl}
				a_{maxdec} \leq V_a u(\tau)  \leq a_{maxacc},
			\end{IEEEeqnarray}
			where $ V_a = \begin{bmatrix}
				1 & 0
			\end{bmatrix}$,
			and $a_{maxdec}$ and $a_{maxacc}$ are the maximum values of the acceleration of the engine and the deceleration of the brake, respectively, within the capability of the vehicle. 
			
			Next, the obstacle avoidance constraint is considered. In this work, the ego vehicle is formulated by a rectangle, and the obstacles (e.g., other vehicles) are formulated as ellipses. As the safe distance at the front and rear of the vehicle is larger, while the safe distance at the sides of the vehicle is smaller, the collision regions can be appropriately represented by ellipses. In this sense, with the  the time stamp $\tau$ and the heading angle of the obstacle $\theta_o$, the rotational matrix is given by
			\begin{IEEEeqnarray*}{rCl} 
				R(\tau)&=&\begin{bmatrix}
					\operatorname{cos}\theta_o(\tau) & 	-\operatorname{sin}\theta_o(\tau) \\
					\operatorname{sin}\theta_o(\tau) & 	 \operatorname{cos}\theta_o(\tau)
				\end{bmatrix}. \yesnumber
			\end{IEEEeqnarray*}
			Then, denote $e_a$ and $e_b$ as the length of semi-major and semi-minor axes corresponding to the ellipse, and $d_{xy}(\tau)$ as the position difference between the ego vehicle and the obstacle, and then referring to~\cite{pan2020safe}, the obstacle avoidance constraint is formulated as
			\begin{IEEEeqnarray*}{rCl} 
				h(x(\tau))&=& 1-d_{xy}(\tau)^{\top} A(\tau)d_{xy}(\tau)\leq 0, \yesnumber
			\end{IEEEeqnarray*}
			with 
			\begin{IEEEeqnarray*}{rCl}
				A(\tau)&=&R(\tau) \begin{bmatrix}
					{\frac{1}{e_a^2}}  & 0\\0 & {\frac{1}{e_b^2}}
				\end{bmatrix}R(\tau)^{\top}. \yesnumber
			\end{IEEEeqnarray*}

			\section{ADMM-based Constrained iLQR Algorithm for Motion Planning}
			In this section, we present a general approach for the motion planning problem. In this following text, we propose the use of a nonlinear system model with $n$ system state variables and $m$ control input variables. Also, for the sake of simplicity, we utilize the vector concatenation symbol, i.e., the notation $a=(a_1,a_2,\dotsm,a_N)$ is equivalent to the notation $a=[a_1^{\top},a_2^{\top},\dotsm,a_N^{\top}]^{\top}$.
			
			\subsection{Motion Planning with iLQR}
			Generally, a motion planning problem in autonomous driving can be formulated as
			\begin{IEEEeqnarray*}{cl}~\label{eq:opt1}
				\displaystyle\operatorname*{minimize}_{\big(x(\tau),u(\tau)\big)\in\mathbb R^{n}\times \mathbb R^{m}}\quad
				&   \phi(x(T))+\sum_{\tau=0}^{T-1}\ell_\tau\big(x(\tau),u(\tau)\big) \\
				\operatorname*{subject\ to}\quad
				&x(\tau+1)=f\big(x(\tau),u(\tau)\big)\\
				&\tau=0,1,\dotsm, T-1\\
				&x(0)=x_0,\yesnumber
			\end{IEEEeqnarray*}
			where $\phi(x(T))$ denotes the terminal cost function to the system state vector in the last time stamp.

			The principle of dynamic programming reduces the minimization over a sequence of control actions to a sequence of minimization over one single control action, which is summarized as the Bellman equation characterized by
			\begin{IEEEeqnarray}{rCl}~\label{eq:Bellman}
				V_\tau\big(x(\tau)\big) = \displaystyle\operatorname*{min}_{u(\tau)} \Big\{\ell_\tau\big(x(\tau),u(\tau)\big)+V_{\tau+1}\Big(f\big(x(\tau),u(\tau)\big)\Big)\Big\},\nonumber\\
			\end{IEEEeqnarray}
			with $V_\tau(x(\tau))$ and $V_{\tau+1}(f(x(\tau),u(\tau)))$ denoting the minimum cost-to-go that starts from $\tau$ and  $\tau+1$, respectively.
			
			For the first step of iLQR, a feasible nominal trajectory $\{\hat u(\tau),\hat x(\tau)\}_{\tau=0}^{T}$ is generated by passing the nominal input trajectory to the nonlinear system with the system initial state variables. Then, the second step is the so-called the backward pass. With the value function at the final time stamp, i.e., $V_T(x(T))=\phi(x(T))$, the iteration proceeds backwards from the time stamp $T-1$ to the first time stamp. 
			
			Consider the right-hand side of \eqref{eq:Bellman}, and we denote the perturbed Q-function as $Q_\tau(\delta x(\tau),\delta u(\tau))$, where $\delta x(\tau)$ and $\delta u(\tau)$ represent the amount of change in terms of the system state vector and control input vector at the time stamp $\tau$ in the two adjacent iterations. The perturbed Q-function can be represented by
			\begin{IEEEeqnarray*}{l} 
				Q_\tau\big(\delta x(\tau),\delta u(\tau)\big)\\
				=\ell_\tau\big({x}(\tau)+\delta{x}(\tau), {u}(\tau)+\delta {u}(\tau)\big)-\ell_\tau\big({x}(\tau), {u}(\tau)\big)\nonumber\\
				+V_{\tau+1}\Big(f\big({x}(\tau)+\delta{x}(\tau), {u}(\tau)+\delta {u}(\tau)\big)\Big)\\
				-V_{\tau+1}\Big(f\big({x}(\tau), {u}(\tau)\big)\Big). \yesnumber
			\end{IEEEeqnarray*}
			Furthermore, by performing second-order Taylor expansion, we approximate the perturbation term by the following matrix equation:
			\begin{IEEEeqnarray}{rCl} 
				&&Q_\tau\big(\delta x(\tau),\delta u(\tau)\big)\nonumber\\
				&&\approx 
				\frac{1}{2}\left[\begin{array}{c}
					1 \\
					\delta x(\tau) \\
					\delta u(\tau)
				\end{array}\right]^{\top}\left[\begin{array}{ccc}
					0 & \left(Q_\tau^{\top}\right)_{x} & \left(Q_\tau^{\top}\right)_{u} \\
					\left(Q_\tau\right)_{x} & \left(Q_\tau\right)_{x x} & \left(Q_\tau\right)_{xu} \\
					\left(Q_\tau\right)_{u} & \left(Q_\tau\right)_{ux} & \left(Q_\tau\right)_{uu}
				\end{array}\right]\left[\begin{array}{c}
					1 \\
					\delta x(\tau) \\
					\delta u(\tau)
				\end{array}\right],\nonumber\\
			\end{IEEEeqnarray}
			with
			\begin{IEEEeqnarray*}{rCl}\label{eq:hes}
				\left(Q_\tau\right)_{x} &=&\left(\ell_{\tau}\right)_x+ f_{x}^{\top}(V_{\tau+1})_x \\
				\left(Q_\tau\right)_{u} &=&\left(\ell_{\tau}\right)_u+ f_{u}^{\top}(V_{\tau+1})_x \\
				\left(Q_\tau\right)_{xx} &=&\left(\ell_{\tau}\right)_{xx}+f_{x}^{\top} (V_{\tau+1})_{xx} f_{x}+(V_{\tau+1})_{x} \cdot f_{xx} \\
				\left(Q_\tau\right)_{ux} &=&\left(\ell_{\tau}\right)_{ux}+f_{u}^{\top} (V_{\tau+1})_{xx} f_{x}+(V_{\tau+1})_{x} \cdot f_{ux} \\
				\left(Q_\tau\right)_{uu} &=&\left(\ell_{\tau}\right)_{uu}+f_{u}^{\top} (V_{\tau+1})_{xx} f_{u}+(V_{\tau+1})_{x} \cdot f_{uu},\yesnumber
			\end{IEEEeqnarray*}
			where the operator $\cdot$ in the last three sub-equations denotes the contraction of a vector with a tensor and the subscript in each matrix function denotes the partial derivative of that function with respect to the notation in the subscript. Recall that the iLQR is a special case of the DDP, and the major difference is that the iLQR uses only the first-order derivative of the dynamic function $f$, so the terms related to $f_{xx}$, $f_{ux}$ and $f_{uu}$ in \eqref{eq:hes} are ignored in our problem. It is also worthwhile to mention that both the iLQR and the DDP use the second-order derivative of the value function.
			
			Hence, to further proceed the backward iterations, the optimal control action for the perturbed Q-function is given by
			\begin{IEEEeqnarray}{rCl} 
				\delta u(\tau)^{*} &=&\operatorname*{argmin}_{\delta u(\tau)} \quad  Q_\tau\big(\delta x(\tau),\delta u(\tau)\big),
			\end{IEEEeqnarray}
			and we have
			\begin{IEEEeqnarray}{rCl} 
				\delta u(\tau)^{*} &=& k(\tau)+K(\tau)\delta x(\tau),
			\end{IEEEeqnarray}
			where $k(\tau)$ and $K(\tau)$ are considered as the feedforward vector and feedback matrix at the time stamp $\tau$. Subsequently, it yields that
			\begin{IEEEeqnarray*}{rCl}
				k(\tau)&=&-\left(Q_\tau\right)_{uu}^{-1} \left(Q_\tau\right)_{u} \IEEEyesnumber\IEEEyessubnumber \label{eq:policy1} \\
				K(\tau)&=&-\left(Q_\tau\right)_{uu}^{-1} \left(Q_\tau\right)_{ux}.\IEEEyessubnumber \label{eq:policy2}
			\end{IEEEeqnarray*}
			Substitute \eqref{eq:policy1} and \eqref{eq:policy2} to the Taylor expansion of perturbed Q-function, and then the difference, gradient, and Hessian of the value function are given by
			\begin{IEEEeqnarray*}{rCl} 
				\Delta V &=&-\frac{1}{2} {k(\tau)}^{\top} \left(Q_\tau\right)_{uu} {k(\tau)} \IEEEyesnumber\IEEEyessubnumber \label{eq:policy3}\\
				(V_\tau)_x &=&\left(Q_\tau\right)_{x}-{K(\tau)}^{\top} \left(Q_\tau\right)_{uu} {k(\tau)} \IEEEyessubnumber\label{eq:policy4}\\
				(V_\tau)_{xx} &=&\left(Q_\tau\right)_{xx}-{K(\tau)}^{\top} \left(Q_\tau\right)_{uu} {K(\tau)}. \IEEEyessubnumber\label{eq:policy5}
			\end{IEEEeqnarray*}
			As follows, the backward pass is recursively conducted by calculating \eqref{eq:policy1}-\eqref{eq:policy2} and \eqref{eq:policy4}-\eqref{eq:policy5}, until the first time stamp is reached.
			
			Then, after the backward pass is completed, the third step is carried out, which is the so-called the forward pass. In the forward pass, the actual trajectory will be generated using the system dynamics and the control action pre-determined by \eqref{eq:policy1} and \eqref{eq:policy2}. Here, we have
			\begin{IEEEeqnarray}{rCl} \label{eq:iLQR_forward_pass}
				u(\tau) &=&\hat{u}(\tau)+k(\tau)+K(\tau)\big(x(\tau)-\hat{x}(\tau)\big)\nonumber \\
				x(\tau+1) &=&f\big(x(\tau), u(\tau)\big).
			\end{IEEEeqnarray}
			Remarkably, the basic idea of the iLQR is to find the optimal control action $\delta u(\tau)^*$ in the backward pass to drive the system, such that the nominal trajectory $(\hat x,\hat u)$ can be updated to a new feasible trajectory in the forward pass. In the end, it will converge to the optimal trajectory asymptotically.

			\subsection{ADMM Algorithm for Constrained iLQR}\label{section:admm}
			If inequality constraints are considered in the motion planning problem, the constrained optimization problem is given by
			\begin{IEEEeqnarray*}{cl}
				\displaystyle\operatorname*{minimize}_{\big(x(\tau),u(\tau)\big)\in\mathbb R^{n}\times \mathbb R^{m}}\quad
				&   \phi(x(T))+\sum_{\tau=0}^{T-1}\ell_\tau\big(x(\tau),u(\tau)\big)\\
				\operatorname{subject\ to} \quad
				& x(\tau+1)=f\big(x(\tau),u(\tau)\big)\\
				& g(y)\ge0\\
				& \tau=0,1,\dotsm,T-1,\yesnumber
			\end{IEEEeqnarray*}
			where the vector $y$ is defined as
			\begin{IEEEeqnarray*}{rl}
				y=\big(u(0),x(1),u(1),\dotsm,x(T)\big)\in\mathbb R^{(n+m)T},
			\end{IEEEeqnarray*}
			and the function $g$ denotes the inequality constraints included in the optimization problem. Notably, the symbol $\geq$ represents the generalized ``greater than or equal to'' in this work. For instance, in the vector case, the symbol $\geq$ denotes the element-wisely ``greater than or equal to''; in the matrix case, the symbol $\geq$ represents the positive semi-definiteness.
			
			To further simplify the optimization problem, we introduce the definition of the indicator function.
			\begin{definition}
				The indicator function with respect to a set $\mathbb B$ is defined as
				\begin{IEEEeqnarray*}{l}
					\delta_\mathbb B(B)=\left\{\begin{array}{ll} 0 & \text{if } B\in \mathbb B\\
						\infty & \text{otherwise}.
					\end{array}\right.\yesnumber
				\end{IEEEeqnarray*}
			\end{definition}
			
			We further define the sets 
			\begin{IEEEeqnarray*}{rCl}
				\mathscr F&=&\Big\{y=\big(u(0),x(1),u(1),\dotsm,x(T)\big)\,\Big|\, \\
				&&x(\tau+1)=f\big(x(\tau),u(\tau)\big),\forall \tau=0,1,\dotsm,T-1\Big\}\\
				\tilde{ \mathscr G}&=&\Big\{y\in\mathbb R^{(n+m)T}\,\Big|\, g(y)\ge0\Big\}.\yesnumber
			\end{IEEEeqnarray*}
			Then the optimization problem can be equivalently denoted by
			\begin{IEEEeqnarray*}{rl}~\label{eq:zl}
				\displaystyle\operatorname*{minimize}_{\big(x(\tau),u(\tau)\big)\in\mathbb R^{n}\times \mathbb R^{m}} \quad
				& \phi(x(T))+\sum_{\tau=0}^{T-1}\ell_\tau\big(x(\tau),u(\tau)\big)\\
				&+\delta_{\mathscr F}\left(y\right)+\delta_{\tilde{\mathscr G}}\left(y\right).\yesnumber\IEEEeqnarraynumspace
			\end{IEEEeqnarray*}
			By introducing the consensus variable $\tilde z$, \eqref{eq:zl} can be further expressed as
			\begin{IEEEeqnarray*}{cl}
				\displaystyle\operatorname*{minimize}_{\big(x(\tau),u(\tau)\big)\in\mathbb R^{n}\times \mathbb R^{m}} \quad
				& \phi(x(T))+\sum_{\tau=0}^{T-1}\ell_\tau\big(x(\tau),u(\tau)\big)\\
				&+\delta_{\mathscr F}\left(y\right)+\delta_{\tilde{\mathscr G}}\left(\tilde z\right)\\
				\operatorname{subject\ to} \quad
				& y-\tilde z=0.\yesnumber
			\end{IEEEeqnarray*}
			Because the optimization variables in the vector $\tilde z$ in terms of the inequality constraints are not exactly related to each variable in the vector $y$, we introduce the operator $\mathcal A:\mathbb R^{(n+m)T}\rightarrow \mathbb R^{pT}$ to extract the variable that is related to the inequality constraints to accelerate the convergence.
			
			Then the optimization problem is converted to 
			\begin{IEEEeqnarray*}{cl}
				\displaystyle\operatorname*{minimize}_{\big(x(\tau),u(\tau)\big)\in\mathbb R^{n}\times \mathbb R^{m}} \quad
				& \phi(x(T))+\sum_{\tau=0}^{T-1}\ell_\tau\big(x(\tau),u(\tau)\big)\\
				&+\delta_{\mathscr F}\left(y\right)+\delta_{\mathscr G}\left(z\right)\\
				\operatorname{subject\ to} \quad
				& \mathcal Ay-z=0,\yesnumber
			\end{IEEEeqnarray*}
			where $z\in\mathbb R^p$ is the new variable that is related to the inequality constraint; $p$ is the number of optimization variables related to the inequality constraints; the new set $\mathscr G$ is represented by
			\begin{IEEEeqnarray}{rl}
				\mathscr G=\{z=\mathcal A\tilde z\,|\, \mathcal G(\tilde z)\ge 0\},
			\end{IEEEeqnarray}
			where $\mathcal G$ is a function of $\tilde z$ that is used for constructing the inequality constraint in the optimization problem.
			
			Define the augmented Lagrangian function of the optimization problem as
			\begin{IEEEeqnarray*}{rCl}
				\mathcal L_\sigma(y,z;\lambda)&=&\phi(x(T))+\sum_{\tau=0}^{T-1}\ell_\tau\big(x(\tau),u(\tau)\big)+\delta_{\mathscr F}\left(y\right)\\
				&&+\delta_{\mathscr G}\left(z\right)+\frac{\sigma}{2}\|\mathcal A y-z+\sigma^{-1}\lambda\|^2-\frac{1}{2\sigma}\|\lambda\|^2,\\\yesnumber
			\end{IEEEeqnarray*}
			where $\sigma\in\mathbb R$ is the penalty parameter of the  augmented Lagrangian function, and $\lambda\in\mathbb R^p$ is the Lagrange multiplier. Following~\cite{boyd2011distributed}, the ADMM iterations are listed as  
			\begin{IEEEeqnarray*}{rCl}
				y^{k+1} &=& \operatorname*{argmin}_y\,\mathcal L_\sigma\Big(y, z^k;\lambda^k\Big)\\
				z^{k+1} &=& \operatorname*{argmin}_z\,\mathcal L_\sigma\Big(y^{k+1}, z;\lambda^k\Big)\\
				\lambda^{k+1}&=&\lambda^k+\sigma\Big(\mathcal Ay^{k+1}-z^{k+1}\Big),\yesnumber
			\end{IEEEeqnarray*}
			where the superscript $(\cdot)^k$ denotes the iteration number in the ADMM algorithm.
			\subsubsection{First ADMM Iteration}
			
			In the first ADMM iteration, it aims at solving the following sub-problem:
			\begin{IEEEeqnarray*}{cl}\label{eq:ADMM_iter1}
				\displaystyle\operatorname*{minimize}_{\big(x(\tau),u(\tau)\big)\in\mathbb R^{n}\times \mathbb R^{m}}\quad
				& \phi(x(T))+\sum_{\tau=0}^{T-1}\ell_\tau\big(x(\tau),u(\tau)\big)\\
				&+\frac{\sigma}{2}\|\mathcal Ay-z^k+\sigma^{-1}\lambda^k\|^2\\
				\text{subject to}\quad 
				& x(\tau+1)=f\big(x(\tau),u(\tau)\big)\\
				& \tau=0,1,\dotsm,T-1.\yesnumber
			\end{IEEEeqnarray*}
			Furthermore, to facilitate the deployment of the iLQR method, the optimization problem~\eqref{eq:ADMM_iter1} can be equivalently expressed as
			\begin{IEEEeqnarray*}{l}\label{eq:ADMM_final_iter1}
				\displaystyle\operatorname*{minimize}_{\big(x(\tau),u(\tau)\big)\in\mathbb R^{n}\times \mathbb R^{m}} \quad
				\phi(x(T))+\displaystyle\sum_{\tau =0}^{T-1}\Bigg(\ell_\tau \big(x(\tau),u(\tau)\big)\\
				\qquad\qquad+\dfrac{\sigma}{2}\Bigg\|\hat {\mathcal A}(\tau)\begin{bmatrix}x(\tau)\\ u(\tau)\end{bmatrix}-\begin{bmatrix}z^k_x(\tau)\\ z^k_u(\tau)\end{bmatrix}+\sigma^{-1}\begin{bmatrix}\lambda^k_x(\tau)\\ \lambda^k_u(\tau)\end{bmatrix}\Bigg\|^2\Bigg)\\
				\qquad\operatorname{subject\ to} \,\quad\quad
				x(\tau+1)=f\big(x(\tau),u(\tau)\big)\\
				\,\quad\qquad\qquad\qquad\quad\,\,\,\,\tau=0,1,\dotsm,T-1,\yesnumber
			\end{IEEEeqnarray*}
			where the operator $\hat {\mathcal A}(\tau):\mathbb R^{n+m}\rightarrow \mathbb R^{p}$ extracts the optimization variable that is related to the inequality constraints in the time stamp $\tau$. 
			\begin{remark}
				Since only dynamic constraints are included in the optimization problem, the optimization problem~\eqref{eq:ADMM_final_iter1} can be solved by the iLQR algorithm efficiently. 
			\end{remark}

			\subsubsection{Second ADMM Iteration}
			
			In the second iteration, it aims at solving the following sub-problem:
			\begin{IEEEeqnarray*}{rl}
				\displaystyle\operatorname*{minimize}_{z\in\mathbb R^{pT}} \quad
				&\|\mathcal A y^{k+1}+\sigma^{-1}\lambda^k-z\|^2\\
				\operatorname{subject\ to} \quad
				& z\in\mathscr G.\yesnumber
			\end{IEEEeqnarray*}
			Then this problem can be separated into $T+1$ sub-problems, which can be solved in a parallel manner. That is, for all $\tau=0,1,\dotsm,T-1$,
			\begin{IEEEeqnarray*}{rl}\label{eq:ADMM_final_iter2}
				\displaystyle\operatorname*{minimize}_{z(\tau)\in\mathbb R^{p}} \quad
				&\bigg\|\hat {\mathcal A}(\tau)\begin{bmatrix}x^{k+1}(\tau)\\u^{k+1}(\tau) \end{bmatrix}+\sigma^{-1}\lambda^k(\tau)-z(\tau)\bigg\|^2\\
				\text{subject to} \quad  
				& g_\tau\big(z(\tau)\big)\ge0,\yesnumber
			\end{IEEEeqnarray*}
			where 
			\begin{IEEEeqnarray*}{rCl}
				z&=&\big(z(0),z(1),\dotsm,z(T-1)\big)\\
				\lambda&=&\big(\lambda(0), \lambda(1), \dotsm, \lambda(T-1)\big).\yesnumber
			\end{IEEEeqnarray*}
			\begin{remark}
				In each time stamp, the sub-problem~\eqref{eq:ADMM_final_iter2} represents a non-convex projection optimization problem. This is because that, for the Euclidean norm constraint, the projection inwards the ellipse is convex, while the projection outwards the ellipse is non-convex. In other words, $g_\tau(z(\tau))\leq 0 $ is convex, while $g_\tau(z(\tau))\geq 0 $ is non-convex. The surrounding vehicles in this work are formulated as ellipses, and the projection problem onto an ellipse can be easily solved with many existing methods in the literature, so the obstacle avoidance constraint can be addressed appropriately even though it is non-convex. It is also worthwhile to mention that the box constraints are convex, but it can still be addressed with a similar projection method in the second ADMM iteration.
			\end{remark}
			
			\subsubsection{Lagrange Multiplier Update}
			Next, the Lagrange multiplier can be updated with the following rule: 
			\begin{IEEEeqnarray}{rl}\label{eq:update_multiplier}
				\lambda^{k+1}=\lambda^k+\sigma\Big(\mathcal Ay^{k+1}-z^{k+1}\Big).
			\end{IEEEeqnarray}
			
			\begin{remark}
				Due to the properties of the ADMM, it circumvents the feasibility requirement of the trajectory at the first iteration.
			\end{remark}
			
\begin{figure}[t]
				\centering
				\includegraphics[trim=0 0 0 0, clip, width=1\linewidth]{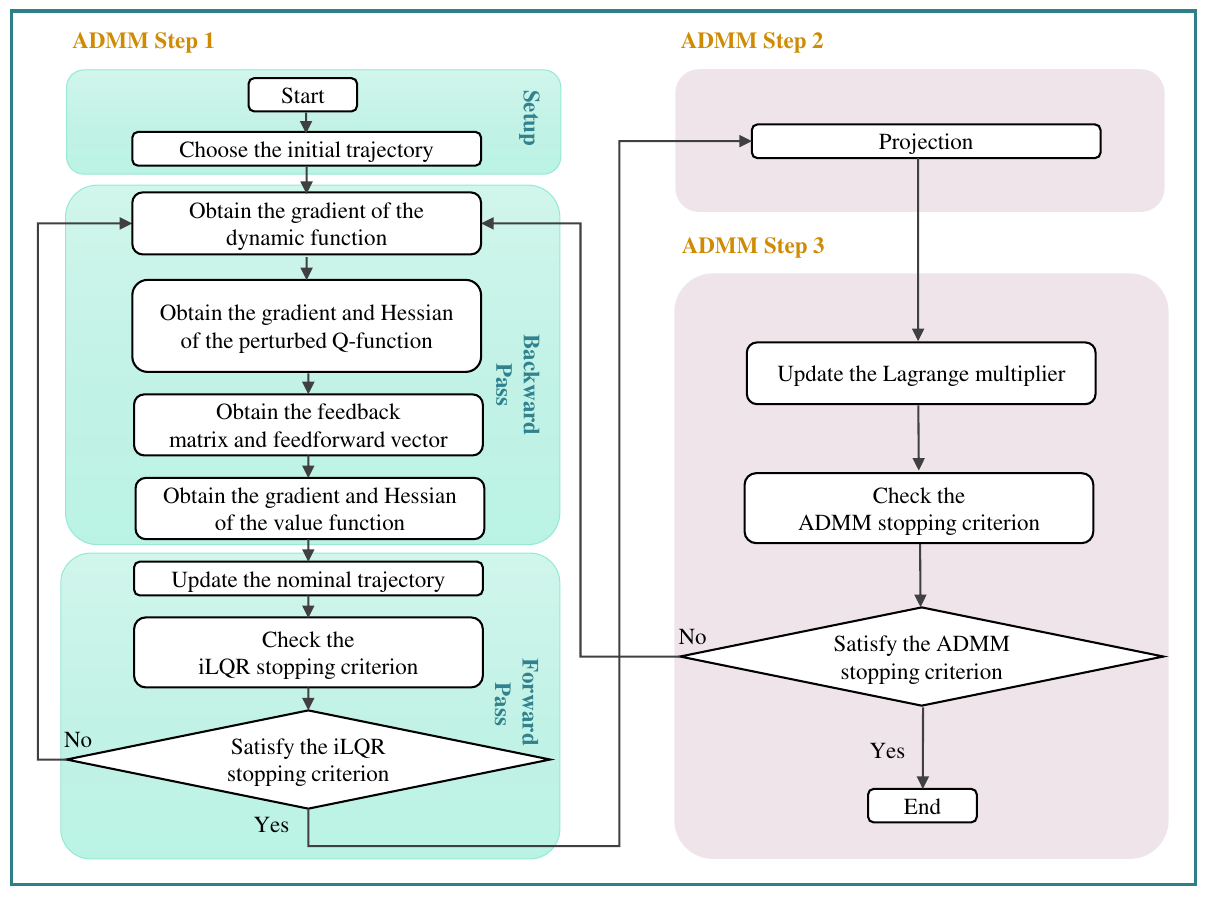}
				\caption{Schematic illustration of the proposed algorithm.}
				\label{fig:ADMMCiLQR4}
			\end{figure}

			To summarize the above discussions, the schematic illustration of the proposed algorithm shown in Fig.~\ref{fig:ADMMCiLQR4}, and the pseudocode is summarized in Algorithm~\ref{algorithm}.
			
			\begin{algorithm}\label{algo}
				\centering
				\caption{ADMM-based Constrained iLQR for Motion Planning}
				\begin{algorithmic}[1]\label{algorithm}
					\REQUIRE
					System dynamic model $f$;\\
					\hspace{9mm}objective function $\phi$ and $\ell$;\\ 
					\hspace{9mm}inequality constraints $g$;\\ 
					\hspace{9mm}initial system state vector $x_0$;\\  
					\hspace{9mm}penalty parameter $\sigma$;\\ 
					\hspace{9mm}the maximum iteration number of ADMM $N^\text{ADMM}_\text{max}$;\\ \hspace{9mm}the maximum iteration number of iLQR $N^\text{iLQR}_\text{max}$. 
					\STATE Generate the initial trajectory $(\hat x(\tau),\hat u(\tau))$.
					\STATE Formulate the optimization problem into the iLQR solvable form by~\eqref{eq:ADMM_final_iter1}.
					\FOR {$i=1,2,\dotsm,N^\text{ADMM}_\text{max}$}
					\STATE \textbf{(ADMM First Iteration)}
					\FOR {$j=1,2,\dotsm,N^\text{iLQR}_\text{max}$}
					\STATE Perform the iLQR backward pass by~\eqref{eq:policy1}-\eqref{eq:policy2} and \eqref{eq:policy4}-\eqref{eq:policy5}.
					\STATE Perform the iLQR forward pass by~\eqref{eq:iLQR_forward_pass}.
					\IF {iLQR stopping criterion is satisfied}
					\STATE \textbf{break}
					\ENDIF
					\ENDFOR
					\STATE \textbf{(ADMM Second Iteration)}
					\STATE Solve the optimization problem~\eqref{eq:ADMM_final_iter2}.
					\STATE \textbf{(Lagrange Multiplier Update)}
					\STATE Update the Lagrange multiplier by~\eqref{eq:update_multiplier}.
					\IF {ADMM stopping criterion is satisfied}
					\STATE \textbf{break}
					\ENDIF
					\ENDFOR
				\end{algorithmic}
			\end{algorithm}
			
			\subsection{Convergence Analysis and Stopping Criterion}
			The ADMM algorithm applied in a convex optimization problem can realize the first-order convergence. However, for non-convex optimization problems, the convergence of the algorithm cannot be guaranteed, yet there are many successful applications of the non-convex ADMM algorithms available in the literature. Besides, the convergence of the non-convex ADMM in a variety of scenarios is also well-established. It is noteworthy to mention that the ADMM algorithm usually shows satisfying convergence in the consensus structure.
			
			Furthermore, the stopping criterion is essential for the algorithm because it will significantly influence the computation efficiency and performance. The stopping criterion for the iLQR iterations is usually chosen as the amount of change of the value function. There are numerous choices of the stopping criterion in terms of the ADMM algorithm, such as the primal-dual residual error, primal residual error, etc. In our case, to simplify the whole optimization process, we use the ADMM iteration number as the stopping criteria. 
			\section{Illustrative Example}

			\begin{figure}[t]
				\centering
				\includegraphics[trim=50 0 100 50, clip, width=0.9\linewidth]{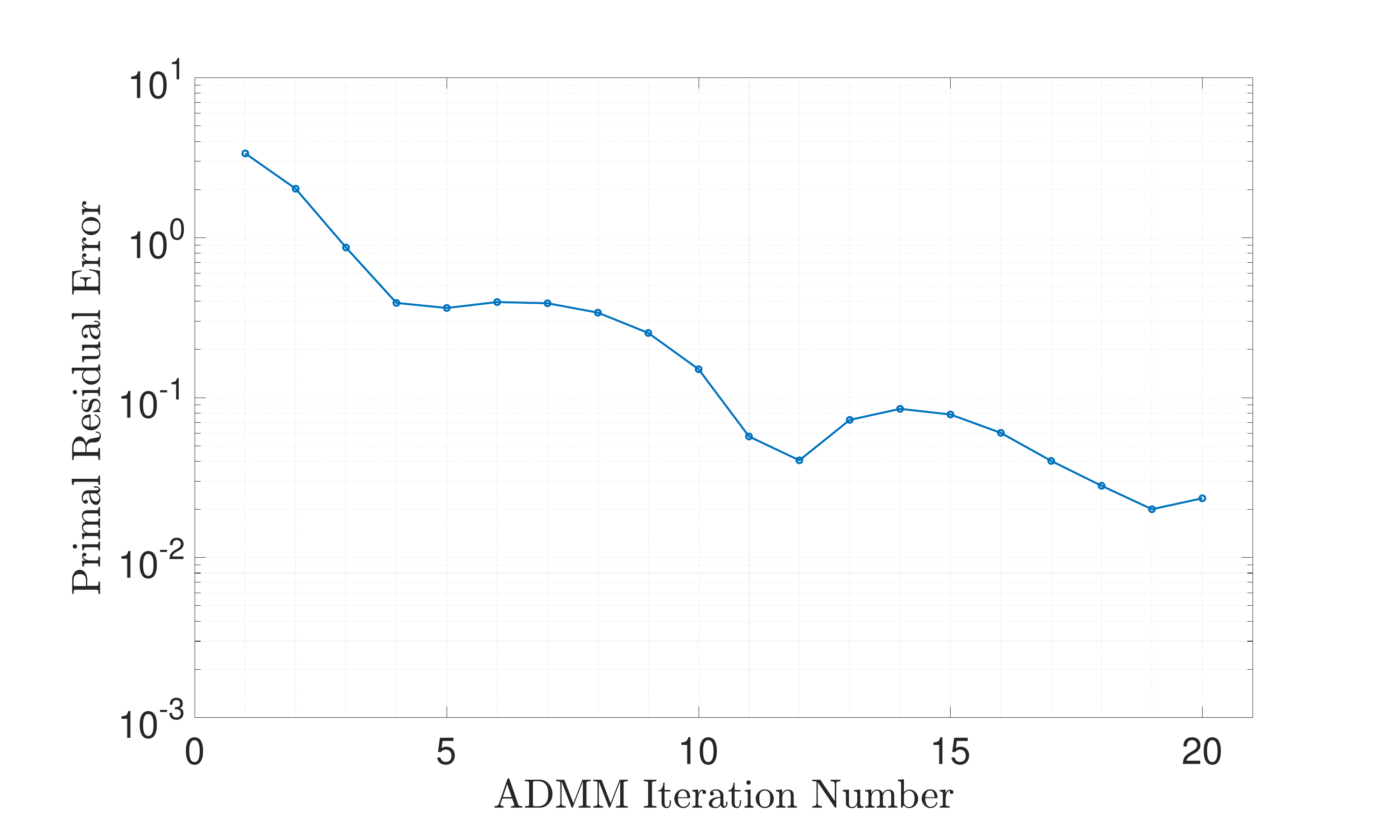}
				\caption{Primal residual error during ADMM iterations in the static obstacle avoidance scenario.}
				\label{fig:Convergence_Static}
			\end{figure}
			
			\begin{figure}[t]
				\centering
				\includegraphics[trim=80 120 100 200, clip, width=0.9\linewidth]{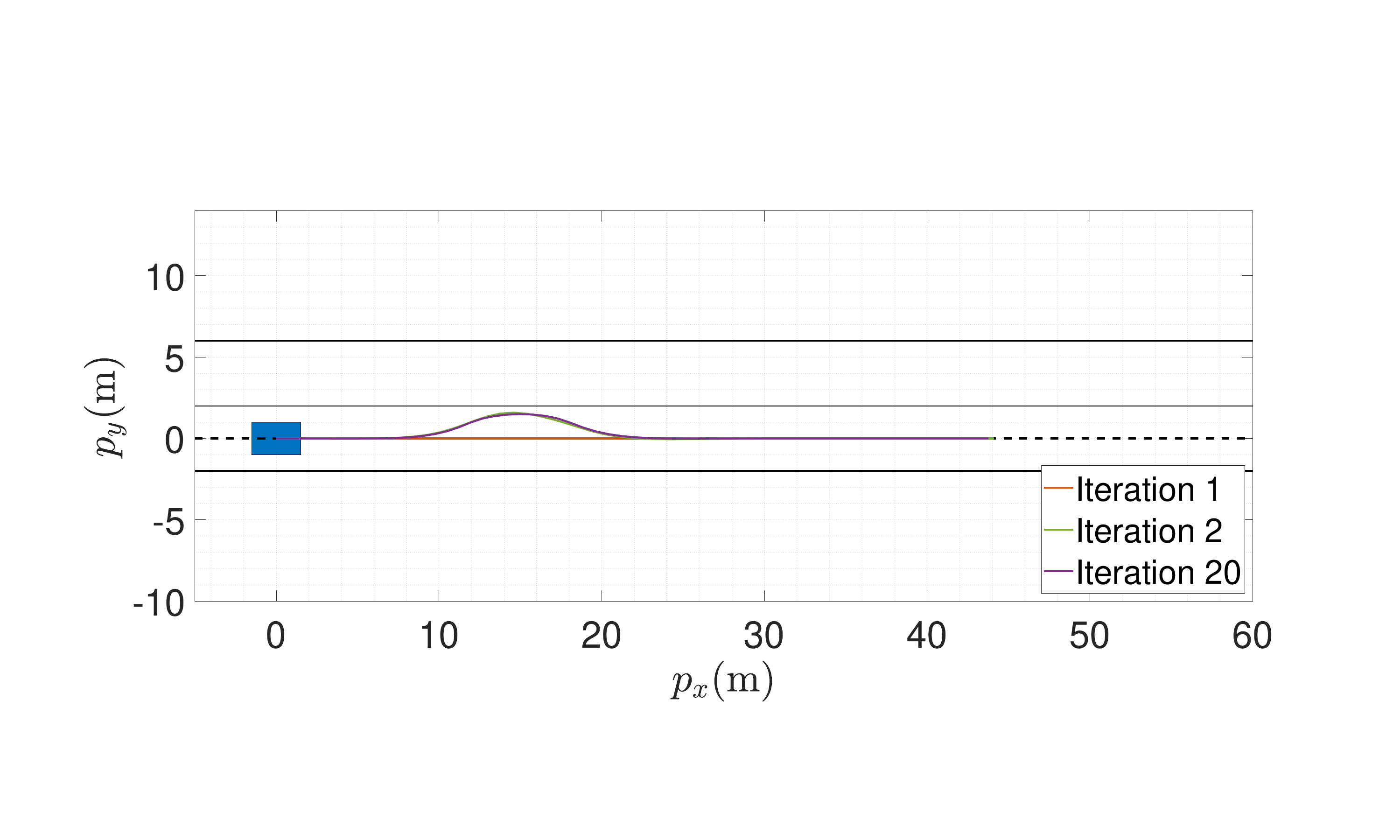}
				\caption{Change of trajectory during iLQR iterations in the static obstacle avoidance scenario.}
				\label{fig:Iteration_Static}
			\end{figure}
			
			\begin{figure}[t]
				\centering
				\includegraphics[trim=80 0 100 100, clip, width=0.9\linewidth]{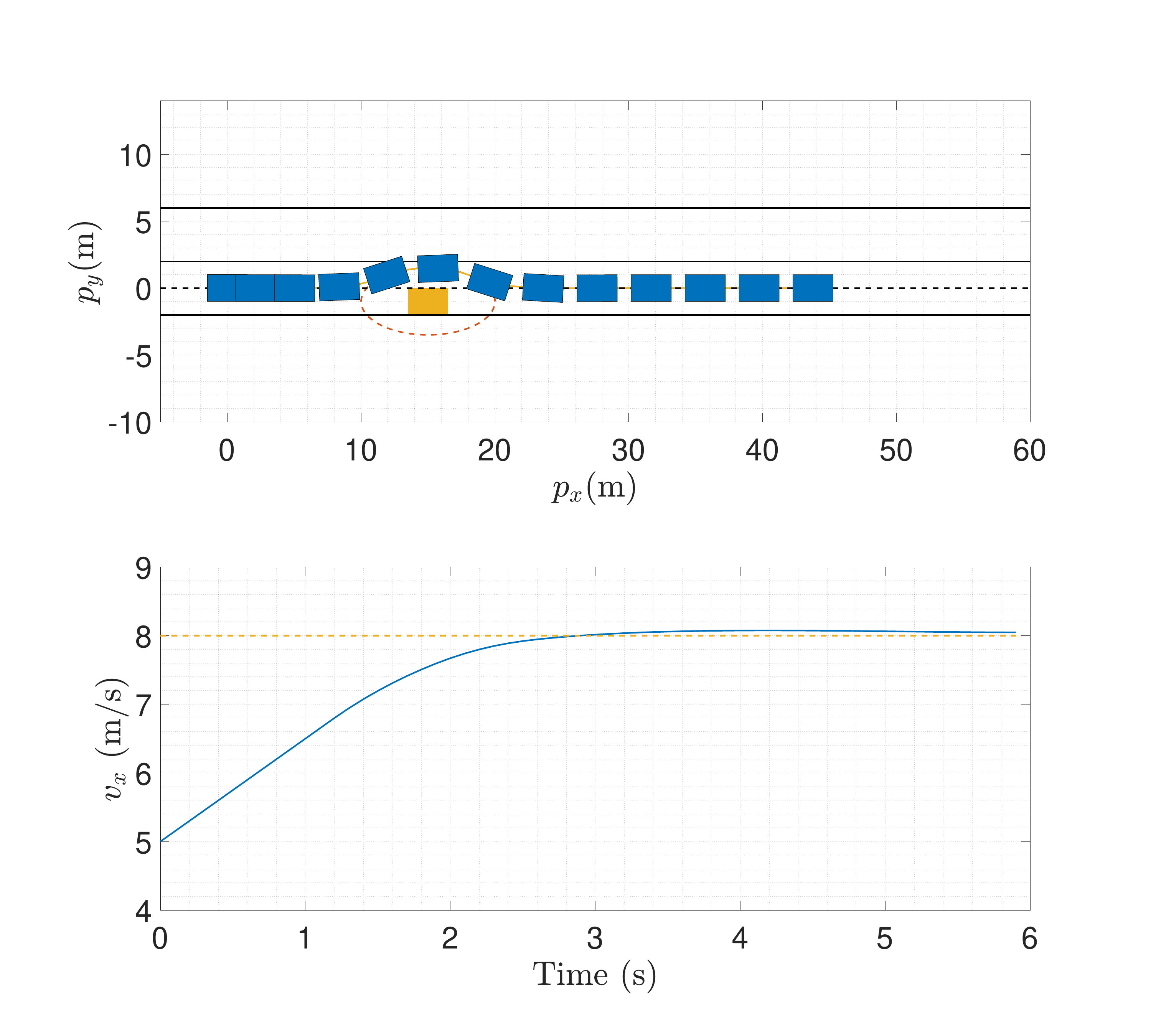}
				\caption{Trajectory and longitudinal velocity of the vehicle in the static obstacle avoidance scenario.}
				\label{fig:Position_Static}
			\end{figure}
			
			\begin{figure}[t]
				\centering
				\includegraphics[trim=50 80 100 100, clip, width=0.9\linewidth]{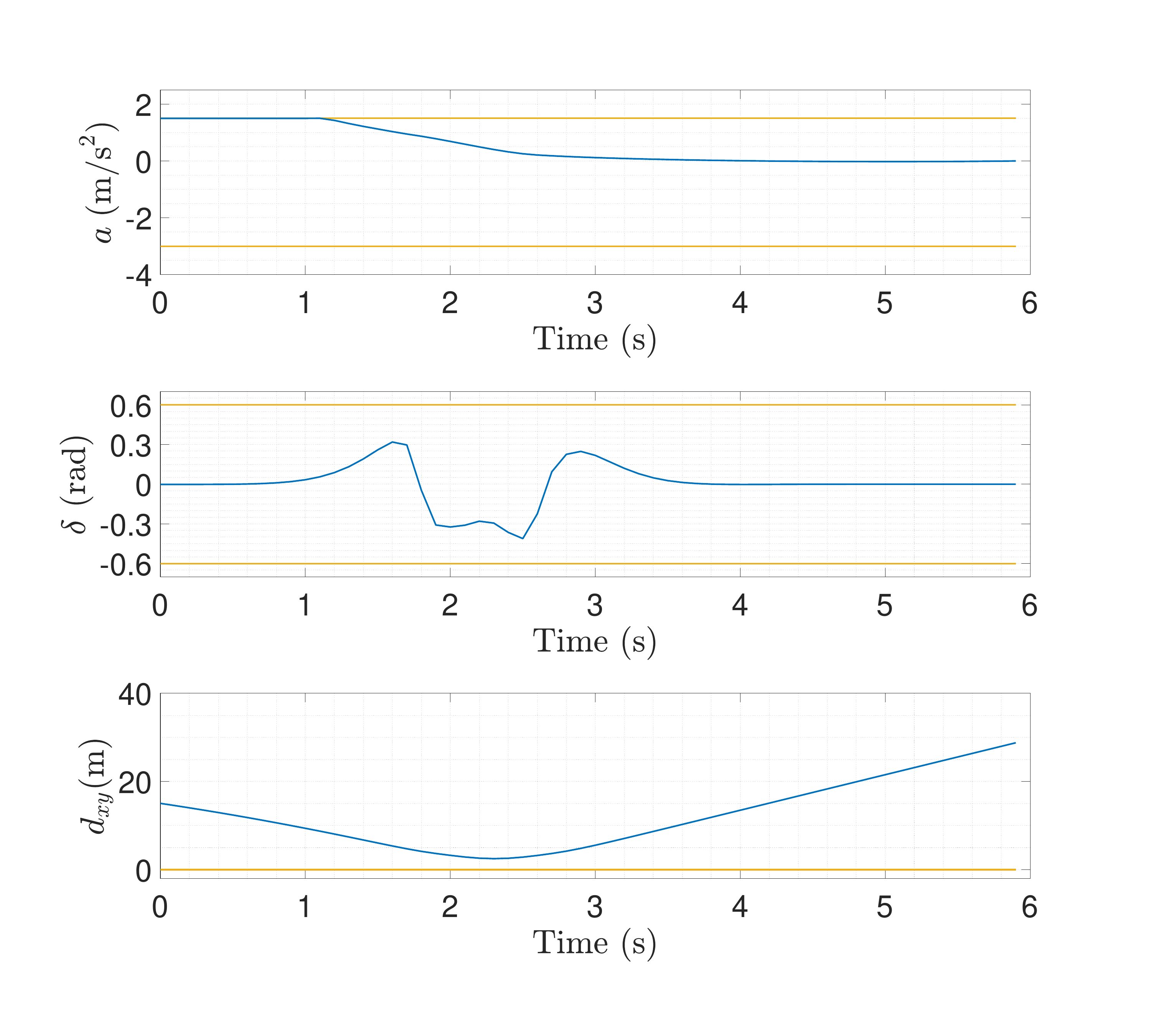}
				\caption{Steering angle and acceleration of the ego vehicle and distance from the ego vehicle to the obstacle in the static obstacle avoidance scenario.}
				\label{fig:Info_Static}
			\end{figure}

			Next, an illustrative example of autonomous driving is presented to validate the effectiveness of the proposed development. Three scenarios are considered here, including one static obstacle avoidance problem and two dynamic obstacle avoidance problems involving lane change and overtaking tasks. The optimization algorithm is implemented in a desktop platform with Intel(R) Xeon(R) W-2225 CPU @ 4.10GHz, the optimization is conducted in Python 3.7 with the Numba package, and the figures are plotted in MATLAB 2020b.    
			
			In the simulation, the solid thick lines in black color represent the road boundaries, the solid thin lines in black color represent the line that separates the road, the dash lines in black color represent the reference of the ego vehicle. Also, the ego vehicle is represented by the blue rectangle, the surrounding vehicles are represented by the yellow rectangle, and collision regions are represented by the ellipse with dash lines. 
		    The mass of the vehicle $m=1412\, \text{kg}$.  The distance from the center of mass to the front and rear axle are $l_f = 1.06 \,\text{m}$ and $l_r= 1.85 \,\text{m}$, respectively.  The cornering stiffness of the front and rear wheels are $k_f=-128916 \,\text{N/rad}$ and $k_r=-85944 \, \text{N/rad}$, respectively. The polar moment of inertia is $I_z = 1536.7 \,\text{kg}\cdot \text{m}^2$. The sampling time is $T_s = 0.1 \,\text{s}$. In all scenarios, the upper bound and lower bound for the steering angle are set as $0.6\, \text{rad}$ and $-0.6\, \text{rad}$, respectively. For the acceleration, the upper bound and lower bound are given by $1.5 \,\text{m}/\text{s}^2$ and $-3\, \text{m}/\text{s}^2$, respectively. The length and width of both the ego vehicle and the surrounding vehicles are $3\, \text{m}$ and $2\, \text{m}$, respectively. In all scenarios, the weightings parameters are set as $q_1=0,q_2=1,q_3=1,r_1=10,r_2=1$ (X-coordinate position tracking error is not penalized in these driving tasks). For the ellipse representing the collision region, the length of semi-major and semi-minor axes is $5\, \text{m}$ and $2.5\, \text{m}$. respectively. The penalty parameter $\sigma$ of the augmented Lagrangian function is set as 10. Also, the prediction horizon is $60$. The maximum iteration numbers of the iLQR algorithm and the ADMM algorithm are set as 100 and 20, respectively.

			\subsection{Scenario 1: Static Obstacle Avoidance }

			In the first scenario, a static obstacle avoidance problem is investigated, where there is one vehicle parked on the street with a coordinate of $(15\,\text{m}, -1\, \text{m})$. For the ego vehicle, the initial position is set as $(0\, \text{m}, 0\, \text{m})$, the initial steering angle is $0 \,\text{rad}$, and the initial longitudinal velocity is $5 \,\text{m}/\text{s}$. The ego vehicle aims to drive along the street and avoid the collision with the vehicle parked on the street. The reference for the position is set as $p_{y}^r=0 \,\text{m}$, and the reference for the longitudinal velocity is $v_x^r=8\,\text{m}/\text{s}$.

			The primal residual error during the ADMM iterations is illustrated in Fig.~\ref{fig:Convergence_Static}. With the proposed approach, the change of trajectory for the ego vehicle during ADMM iterations is plotted in Fig.~\ref{fig:Iteration_Static}. To visualize the change clearly, only the trajectories of the ego vehicle in iteration 1, iteration 2, and iteration 20 are given. From this figure, it shows that a feasible trajectory is obtained finally. With the planned trajectory, the simulation is carried out, and Fig.~\ref{fig:Position_Static} illustrates the trajectory and longitudinal velocity of the ego vehicle. Note that in this scenario, the position of the ego vehicle is plotted every $1\,\text{s}$. As indicated from this figure, the ego vehicle follows the planned trajectory (represented by the solid yellow line in the background) closely, and it successfully avoids the obstacle (the vehicle parked on the street). From the longitudinal velocity profile, it is clear that to avoid this static obstacle, the ego vehicle accelerates at the beginning, and constantly converges to its desired value (represented by the dash yellow line). 
			
			Moreover, to validate the constraint satisfaction conditions, Fig.~\ref{fig:Info_Static} is given, where the steering angle and acceleration of the ego vehicle are depicted. Also, the distance from the ego vehicle to the obstacle (denoted by $d_{xy}$) is presented from this figure. It can be seen that the steering angle and the acceleration are all bounded based on our settings. Besides, the distance between the ego vehicle and the obstacles indicates that no collision occurs in this scenario.
			
			\subsection{Scenario 2: Dynamic Obstacle Avoidance--Lane Change }
			
			In the second scenario, a dynamic obstacle avoidance problem is considered, where a lane change task in autonomous driving is discussed. Under this circumstance, it is assumed that there are two surrounding vehicles. One of them is in front of the ego vehicle, which has an initial position of $(20 \,\text{m}, 0\,\text{m})$ and a constant longitudinal velocity of $3 \, \text{m}/\text{s}$. Also, there is another vehicle on the adjacent lane (the target lane after the lane change). For the vehicle on the target lane, it is initially located at $(0 \,\text{m}, 4\,\text{m})$ and it moves along the street with a longitudinal velocity of $6 \,\text{m}/\text{s}$. Note that the initial position of the ego vehicle is $(0\, \text{m}, 0\, \text{m})$, the initial steering angle is $0 \,\text{rad}$, and the initial longitudinal velocity is $8 \,\text{m}/\text{s}$. The ego vehicle aims to change the lane while avoiding the collisions with the two vehicles driving on the street. The references for the position and the longitudinal velocity are set as $p_{y}^r=4 \,\text{m}$ and $v_{x}^r=8 \,\text{m/s}$.  
			
\begin{figure}[t]
				\centering
				\includegraphics[trim=50 0 100 50, clip, width=0.9\linewidth]{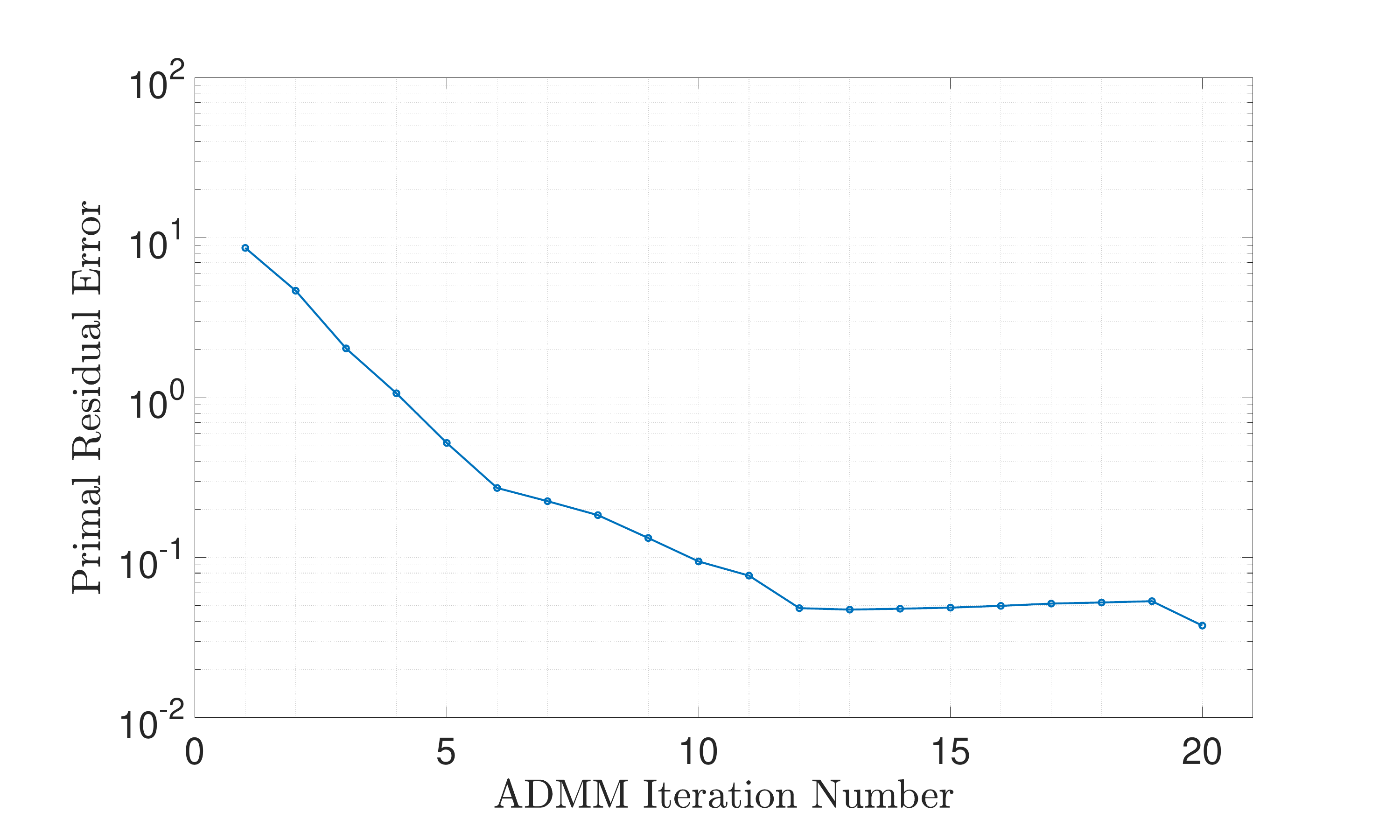}
				\caption{Primal residual error during ADMM iterations in the lane change scenario.}
				\label{fig:Convergence_Lane_Change}
			\end{figure}

\begin{figure}[t]
				\centering
				\includegraphics[trim=80 120 100 200, clip, width=0.9\linewidth]{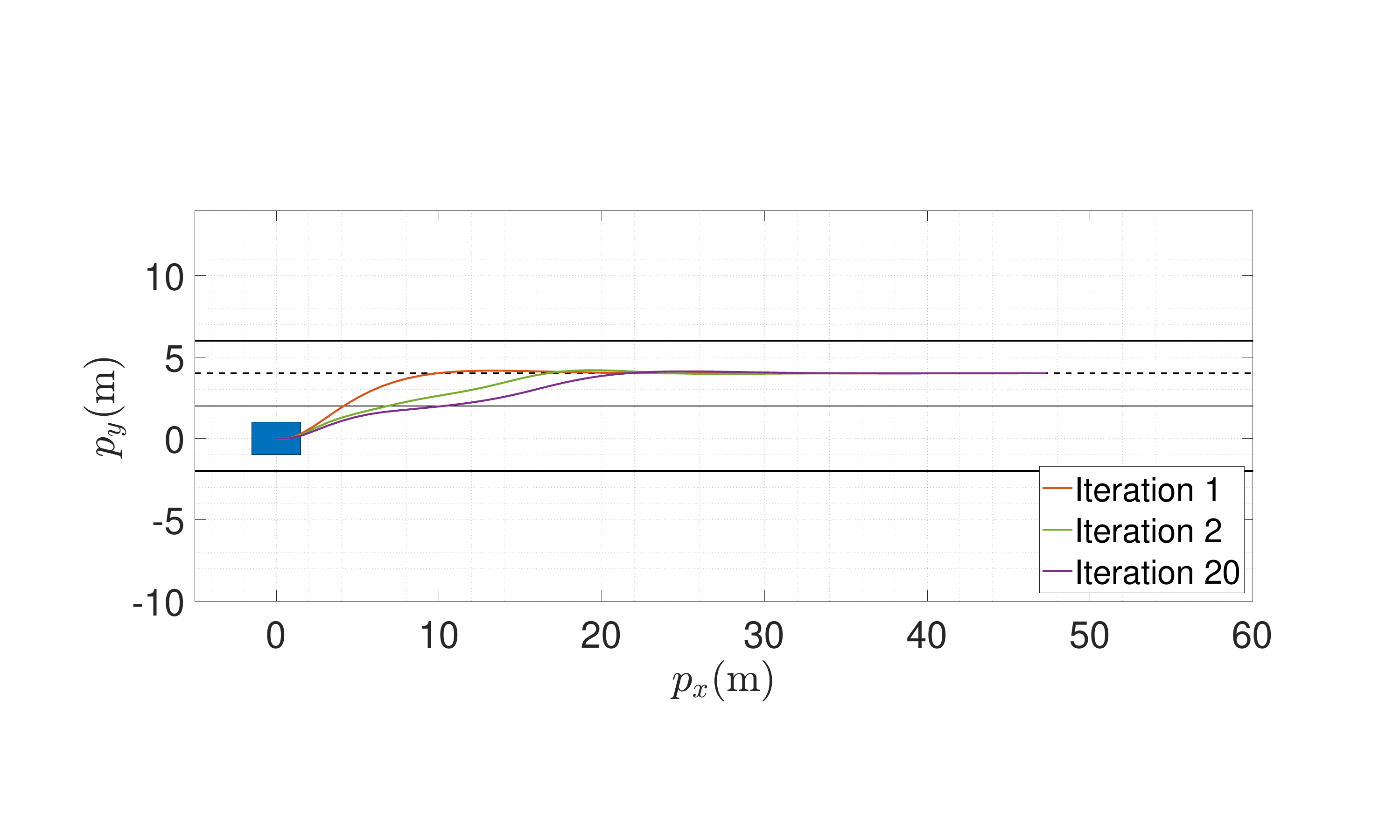}
				\caption{Change of trajectory during iLQR iterations in the lane change scenario.}
				\label{fig:Iteration_Lane_Change}
			\end{figure}

\begin{figure}[t]
				\centering
				\includegraphics[trim=80 0 100 100, clip, width=0.9\linewidth]{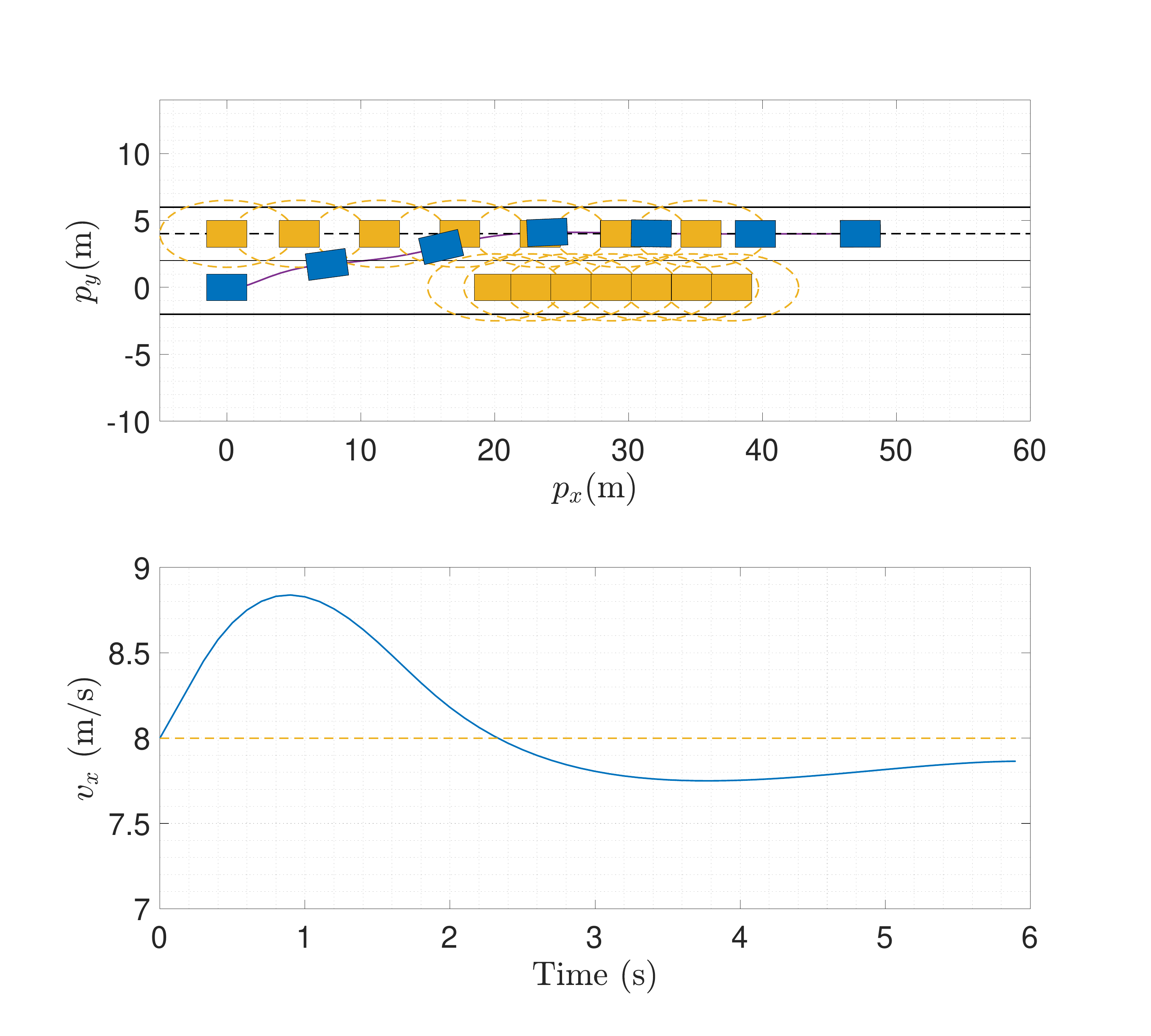}
				\caption{Trajectory and longitudinal velocity of the vehicle in the lane change scenario.}
				\label{fig:Lane_Change}
			\end{figure}

			Similarly, the primal residual error during the ADMM iterations is illustrated in Fig.~\ref{fig:Convergence_Lane_Change}. The change of trajectory for the ego vehicle during ADMM iterations is depicted in Fig.~\ref{fig:Iteration_Lane_Change}, where its trajectories in iteration 1, iteration 2, and iteration 20 are plotted. From this figure, the asymptotic convergence of the trajectory is validated. Subsequently, the trajectory of all the vehicles and the longitudinal velocity of the ego vehicle is given in Fig.~\ref{fig:Lane_Change}. In this case, the position of all the vehicles is plotted every $1 \,\text{s}$. It can be observed from this figure that the ego vehicle follows the planned trajectory (represented by the solid purple line in the background) closely, and no collision with the surrounding vehicles happens. It is worthwhile to mention that, though there are some interactions between the ego vehicle and the collision regions (represented by the ellipses), actually they do not collide at the same time stamp. From the longitudinal velocity profile, it can be seen that the ego vehicle accelerates at the beginning, and converges to around $8 \,\text{m}/\text{s}$. 
			
\begin{figure}[t]
				\centering
				\includegraphics[trim=50 80 100 100, clip, width=0.9\linewidth]{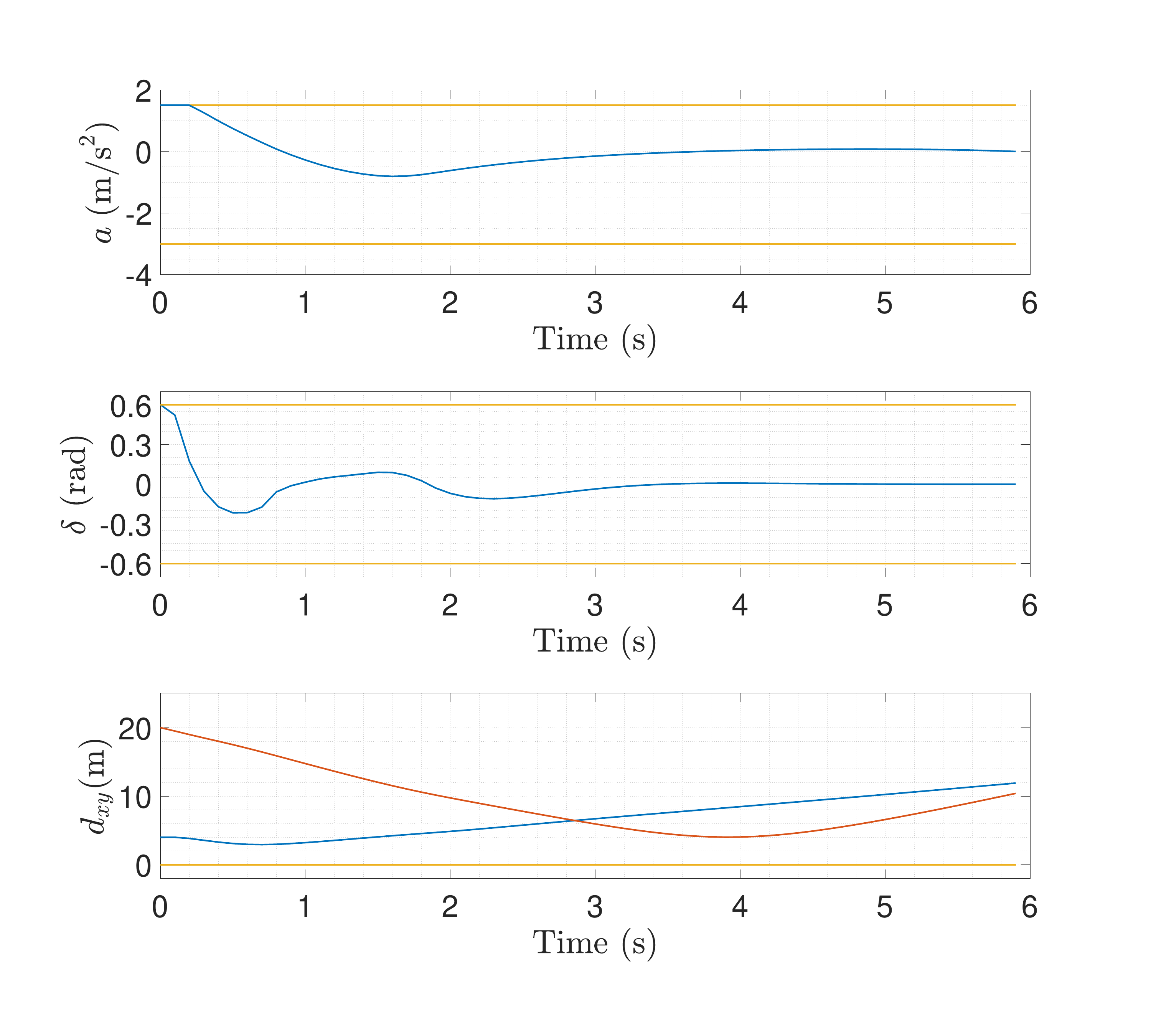}
				\caption{Steering angle and acceleration of the ego vehicle and distance from the ego vehicle to the obstacles in the lane change scenario.}
				\label{fig:Info_Lane_Change}
			\end{figure}

			Additionally, Fig.~\ref{fig:Info_Lane_Change} is shown to validate the constraint satisfaction conditions. As indicated from this figure, the steering angle and acceleration of the ego vehicle are all constrained within the predefined bounds. In terms of the last subplot in Fig.~\ref{fig:Info_Lane_Change}, the curves in the red color and the blue color represent the distance from the ego vehicle to the obstacle on $p_y=0\,\text{m}$  and $p_y=4\,\text{m}$, respectively. The results clearly illustrate that the ego vehicle avoids the surrounding vehicles successfully in this scenario.
			
			
			\subsection{Scenario 3: Dynamic Obstacle Avoidance--Overtaking}
					In the third scenario, an overtaking task is investigated considering two surrounding vehicles. One of them is on the adjacent lane, which has an initial position of $(10 \,\text{m}, 4\,\text{m})$ and an longitudinal velocity of $10 \, \text{m}/\text{s}$. Also, there is another vehicle in the front of the ego vehicle, which is initially located at $(30 \,\text{m}, 0\,\text{m})$ and moves along the street with an initial velocity of $3 \,\text{m}/\text{s}$, and the its velocity  firstly increases to $8\, \text{m/s}$ and then decreases to $3\, \text{m/s}$. Note that the initial position of the ego vehicle is $(0\, \text{m}, 0\, \text{m})$, the initial steering angle is $0 \,\text{rad}$, and the initial longitudinal velocity is $15 \,\text{m}/\text{s}$. The ego vehicle aims to perform the overtaking task and change to the original lane in the end. The reference for the position is set as $p_{y}^r=0 \,\text{m}$, and the reference for the longitudinal velocity is $v_x^r=15\,\text{m}/\text{s}$.
			
\begin{figure}[t]
				\centering
				\includegraphics[trim=50 0 100 50, clip, width=0.9\linewidth]{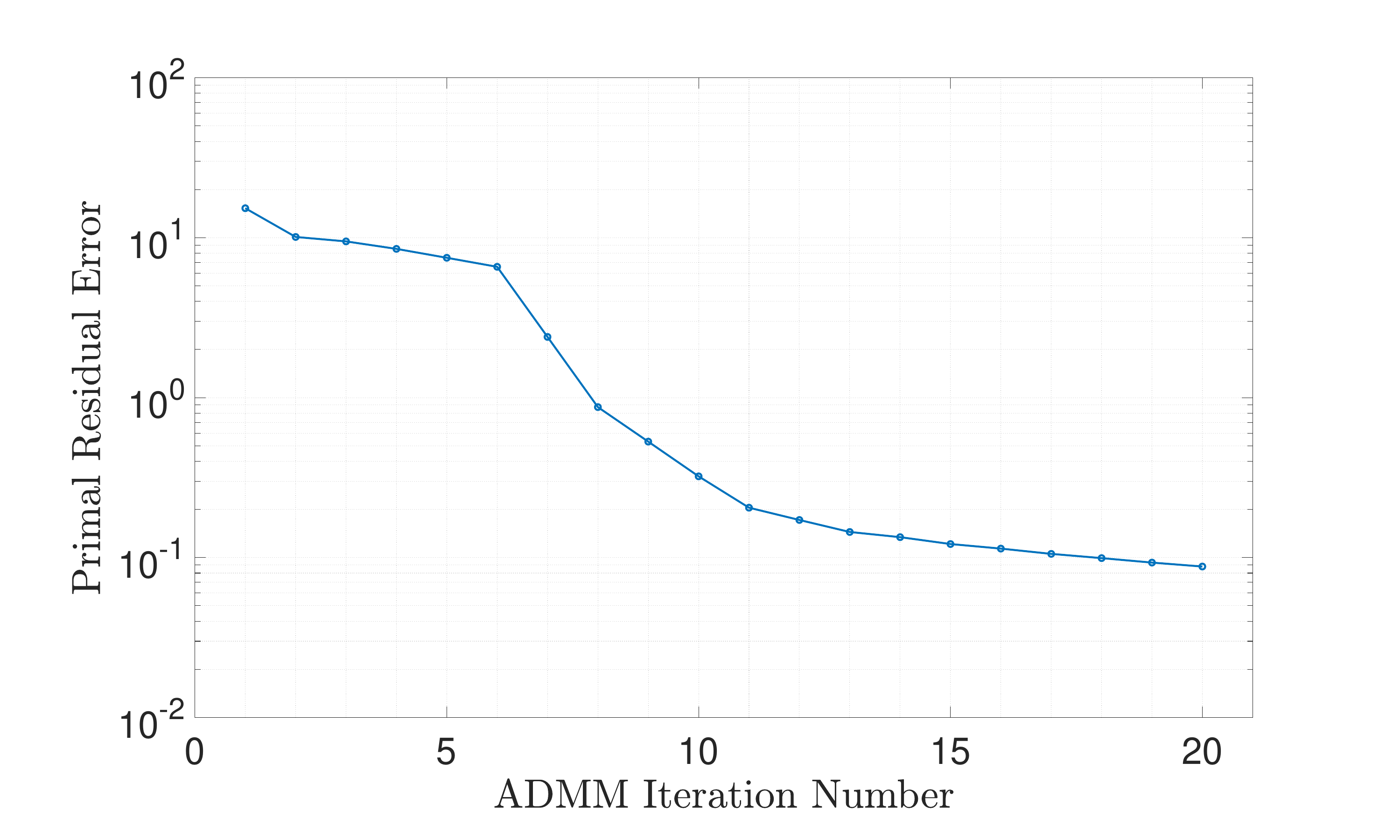}
				\caption{Primal residual error during ADMM iterations in the overtaking scenario.}
				\label{fig:Convergence_Overtaking}
			\end{figure}

\begin{figure}[t]
				\centering
				\includegraphics[trim=80 120 100 200, clip, width=0.9\linewidth]{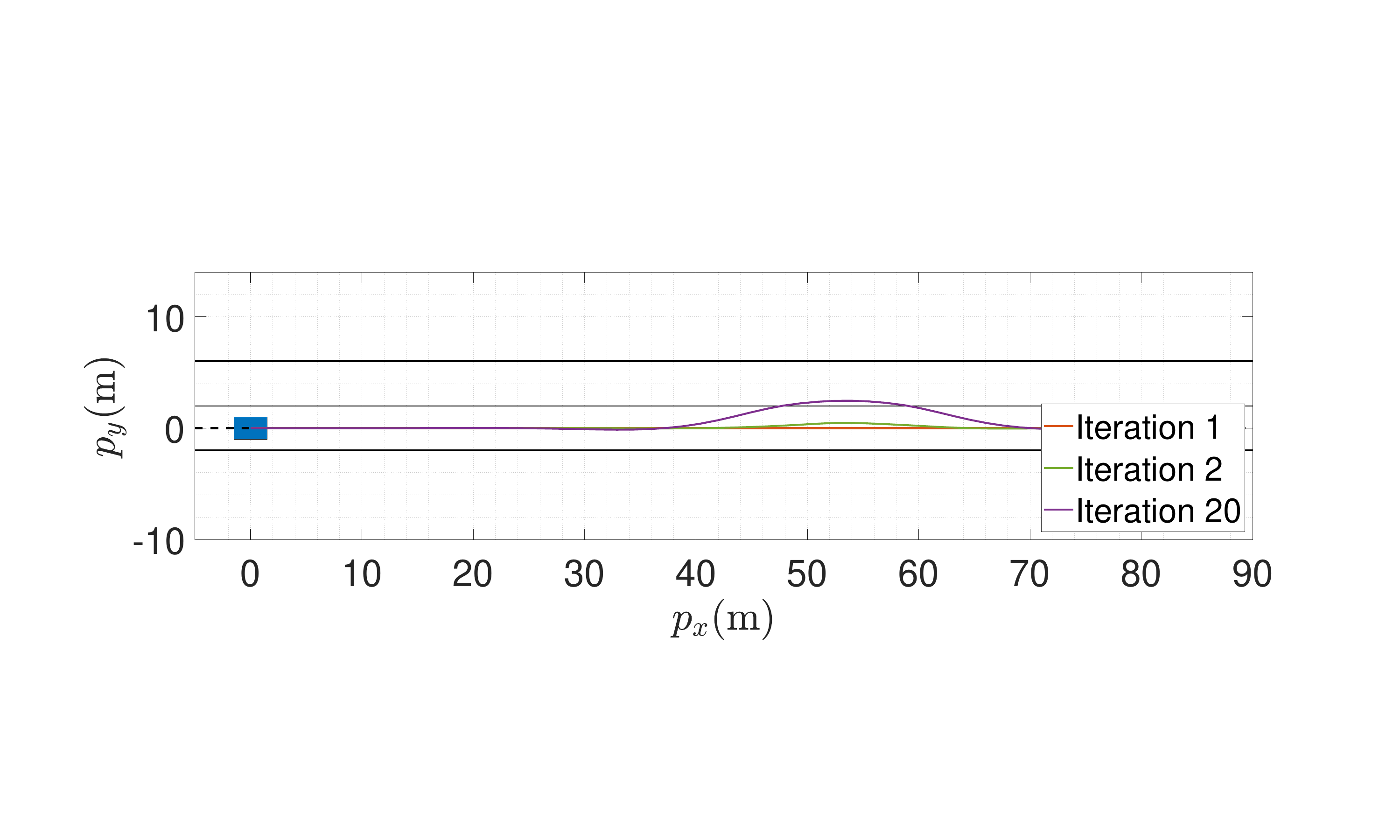}
				\caption{Change of trajectory during iLQR iterations in the overtaking scenario.}
				\label{fig:Iteration_Overtaking}
			\end{figure}

\begin{figure}[t]
				\centering
				\includegraphics[trim=80 0 100 180, clip, width=0.9\linewidth]{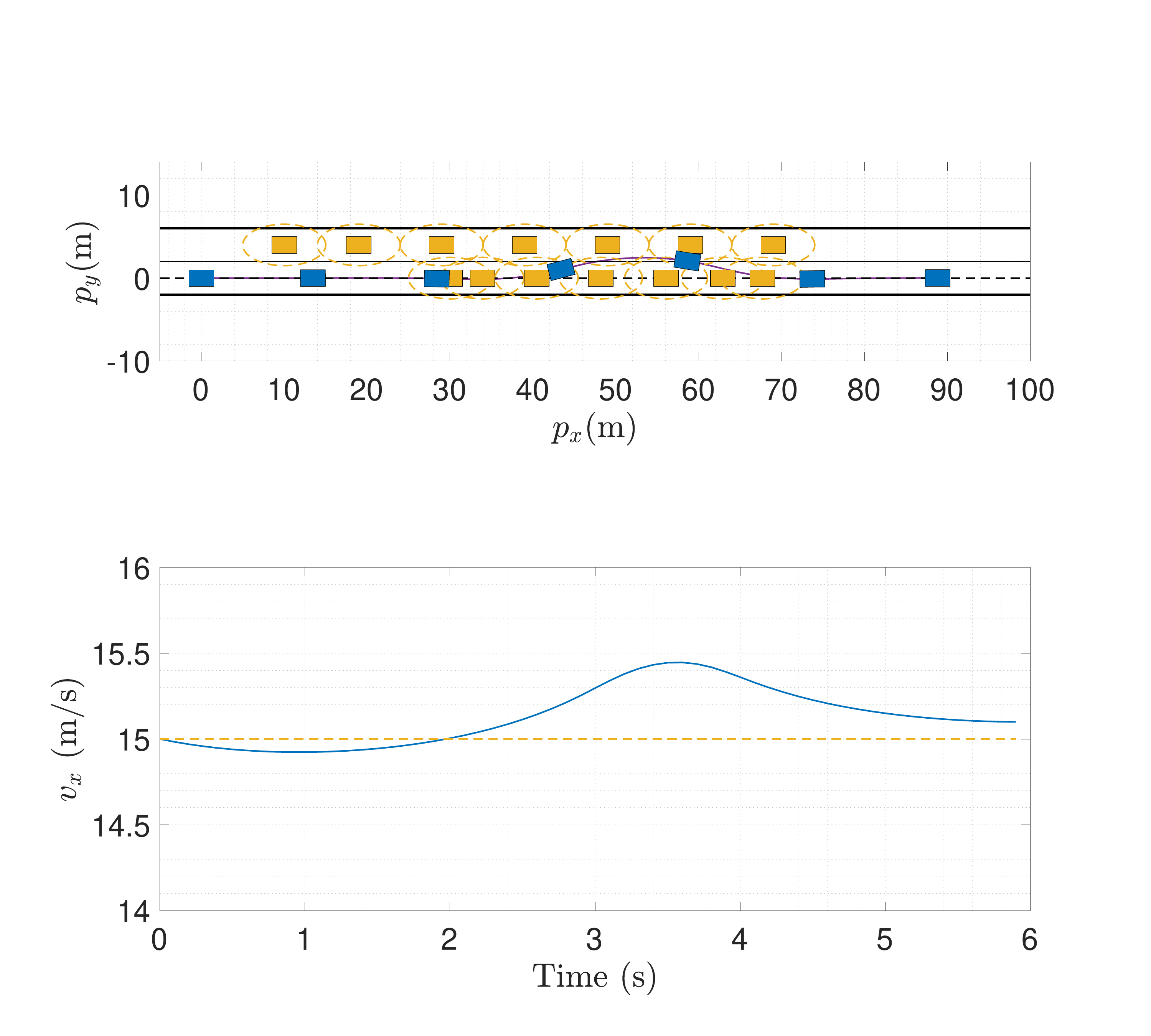}
				\caption{Trajectory and longitudinal velocity of the vehicle in the overtaking scenario.}
				\label{fig:Overtaking}
			\end{figure}

\begin{figure}[t]
				\centering
				\includegraphics[trim=50 80 100 100, clip, width=0.9\linewidth]{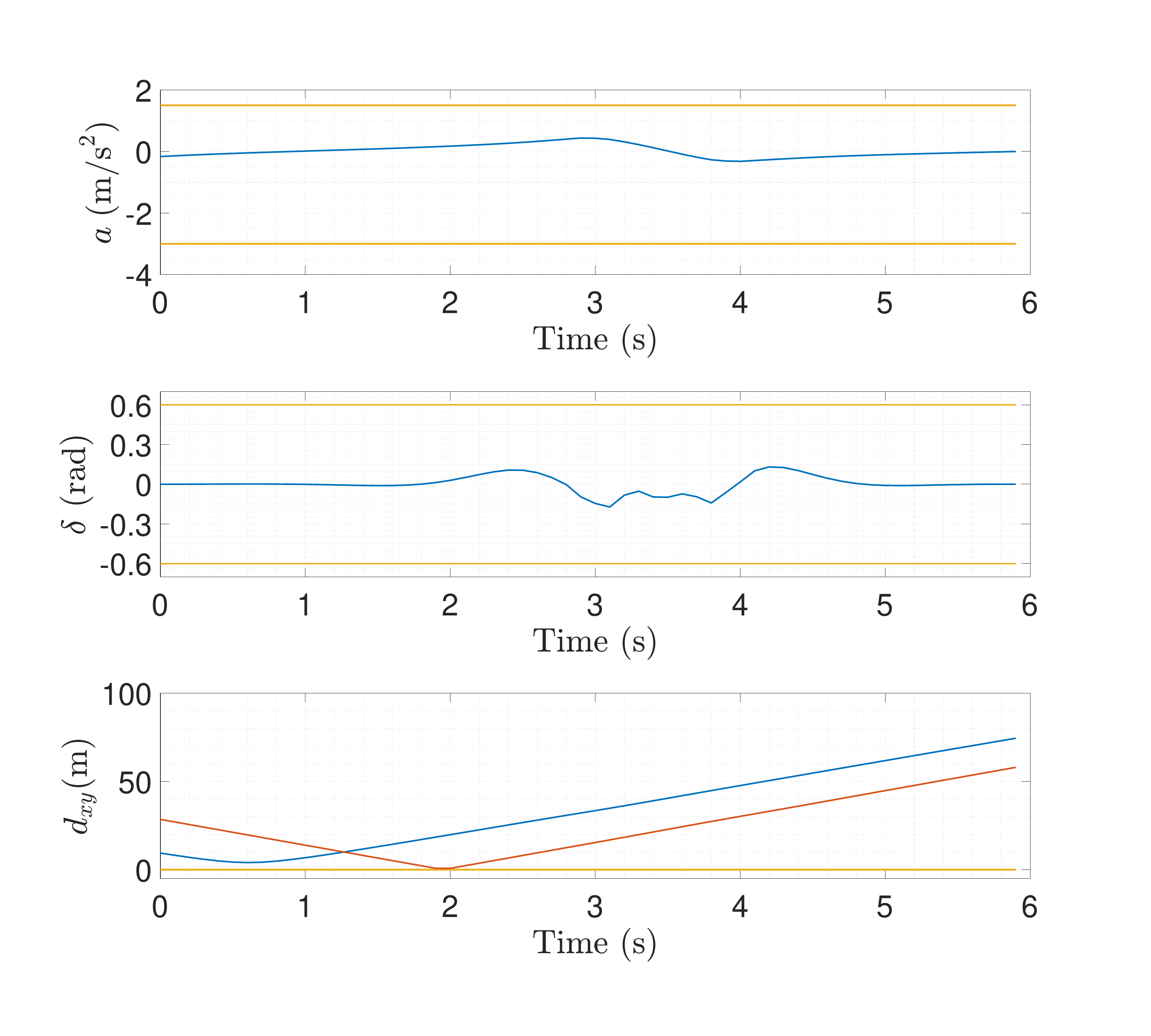}
				\caption{Steering angle and acceleration of the ego vehicle and distance from the ego vehicle to the obstacles in the overtaking scenario.}
				\label{fig:Info_Overtaking}
			\end{figure}

		The primal residual error and the change of trajectory for the ego vehicle during the ADMM iterations are illustrated in Fig.~\ref{fig:Convergence_Overtaking} and Fig.~\ref{fig:Iteration_Overtaking}, respectively. Note that the asymptotic convergence of the trajectory is validated. Also, the trajectories of all the vehicles and the longitudinal velocity of the ego vehicle are given in Fig.~\ref{fig:Overtaking}. Similarly, the position of all the vehicles is plotted every $1 \,\text{s}$. The different line types and line colors have the same meaning in Scenario 2. It can be seen that no collision with the surrounding vehicles happens. Also, from the longitudinal velocity profile, it can be seen that the ego vehicle accelerates at the beginning, and converges to around $15 \,\text{m}/\text{s}$. Similarly, Fig.~\ref{fig:Info_Overtaking} shows that all the constraints are satisfied appropriately, including the bound of the steering angle and acceleration, as well as the distance from the ego vehicle to the surrounding vehicles.  
			
			
			\subsection{Discussion on Computation Time}
			For the optimization in all scenarios, 5 trials with a prediction horizon of $T=60$ are performed, and the computation time by using our proposed approach (denoted by Method 1) is recorded and shown in Table I. The average computation time in these three scenarios is calculated, which is given by $0.0678 \,\text{s}$, $0.1583 \,\text{s}$, and $0.0771 \,\text{s}$, respectively. 
			
			Apart from this, comparison studies are conducted. In particular, Method 2 uses the constrained iLQR algorithm with the logarithmic barrier function~\cite{chen2019autonomous} and Method 3 uses the optimization solver IPOPT~\cite{wachter2006implementation}. In Scenario 1, if the initial longitudinal velocity of the ego vehicle is set as  $5 \,\text{m}/\text{s}$, a feasible trajectory cannot be obtained. Thus, we set the initial longitudinal velocity as  $0 \,\text{m}/\text{s}$. Also, in Scenario 2, if the initial longitudinal velocity of the ego vehicle is set as  $8 \,\text{m}/\text{s}$, a feasible trajectory cannot be obtained because it will collide with the vehicle in the front. Hence, we set the initial longitudinal velocity as  $4 \,\text{m}/\text{s}$ in this case. Similarly, we set the initial longitudinal velocity as  $4 \,\text{m}/\text{s}$ Scenario 3. Subsequently, the computation time is recorded and presented in Table I. Notably, with Method 2, the average computation time in these three scenarios is given by $0.0996 \,\text{s}$, $0.2575 \,\text{s}$, and $0.1391 \,\text{s}$, respectively. Taking Method 2 as the baseline, our proposed method reduces the average computation time by $31.93\%$, $38.52\%$, and $44.57\%$ in the three scenarios, respectively.  With Method 3, the average computation time in these three scenarios is given by $0.1256 \,\text{s}$, $0.3387 \,\text{s}$, and $0.6661 \,\text{s}$, respectively. Taking Method 3 as the baseline, our proposed method reduces the average computation time by $46.02\%$, $53.26\%$, and $88.43\%$ in the three scenarios, respectively. Apparently, when using Method 1 and Method 2, Scenarios 1 and 3 take shorter time for computation, while Scenario 2 takes longer time. However, with Method 3, Scenario 3 takers longer time for the computation.

			\begin{table*}[h]~\label{tab} 
				\centering
				\caption{Computation Time in the All Three Driving Scenarios}
				\begin{tabular}{|c|c|c|c|c|c|c|c|c|c|}
					\hline
					\multirow{2}{*}{} & \multicolumn{3}{c|}{\begin{tabular}[c]{@{}c@{}}Computation Time\\ in Scenario 1\end{tabular}} & \multicolumn{3}{c|}{\begin{tabular}[c]{@{}c@{}}Computation Time\\ in Scenario 2\end{tabular}} &
					\multicolumn{3}{c|}{\begin{tabular}[c]{@{}c@{}}Computation Time\\ in Scenario 3\end{tabular}} \\
					\cline{2-10} 
					& Method 1                                      & Method 2                                      & Method 3
					&Method 1                                      & Method 2      &Method 3           	& Method 1                                      & Method 2   
					& Method 3  \\ \hline
					Trial 1           & $0.0591 \,\text{s}$                           & $0.0910 \,\text{s}$                           & $0.1227 \,\text{s}$                           & $0.1510 \,\text{s}$     
					& $0.2550 \,\text{s}$                           & $0.3399 \,\text{s}$ 
					& $0.0768 \,\text{s}$     
					& $0.1334 \,\text{s}$                           & $0.6657 \,\text{s}$ \\ \hline
					Trial 2           & $0.0549 \,\text{s}$                           & $0.1080 \,\text{s}$                           & $0.1238 \,\text{s}$                           & $0.1786 \,\text{s}$      
					& $0.2614 \,\text{s}$                           & $0.3357\,\text{s}$ 
					& $0.0686 \,\text{s}$     
					& $0.1496 \,\text{s}$                           & $0.6653 \,\text{s}$ \\ \hline
					Trial 3           & $0.0836 \,\text{s}$                           & $0.1074 \,\text{s}$                           & $0.1285 \,\text{s}$                           & $0.1548 \,\text{s}$         
					& $0.2574 \,\text{s}$                           & $0.3413 \,\text{s}$
					& $0.0801 \,\text{s}$     
					& $0.1420 \,\text{s}$                           & $0.6645 \,\text{s}$ \\ \hline
					Trial 4           & $0.0824 \,\text{s}$                           & $0.1102 \,\text{s}$                           & $0.1297 \,\text{s}$                           & $0.1512 \,\text{s}$    
					& $0.2516 \,\text{s}$                           & $0.3368 \,\text{s}$ 
					& $0.0837 \,\text{s}$     
					& $0.1358 \,\text{s}$                           & $0.6712 \,\text{s}$ \\ \hline
					Trial 5           & $0.0591 \,\text{s}$                           & $0.0813 \,\text{s}$                           & $0.1235 \,\text{s}$                           & $0.1558 \,\text{s}$   
					& $0.2619 \,\text{s}$                           & $0.3399 \,\text{s}$ & $0.0763 \,\text{s}$     
					& $0.1346 \,\text{s}$                           & $0.6640 \,\text{s}$ \\ \hline
				\end{tabular}
			\end{table*}
			
			As clearly observed from Table I, the proposed algorithm is highly efficient, which makes it possible to implement in a real-time framework. Compared with the existing works on the constrained iLQR algorithm with the logarithmic barrier function and the optimization solver IPOPT, this work leads to less computation effort. Also, it avoids the feasibility requirement for the trajectory at the first iteration. Notably, the computation efficiency of the algorithm can be further enhanced by choosing a loose stopping criterion. The prediction horizon in the given examples is large enough to realize the on-road driving tasks, and it is also possible to reduce the computation time by choosing a smaller prediction horizon.

			\section{Conclusion}

			This work investigates the motion planning problem in autonomous driving applications. Considering the nonlinear dynamics of the vehicle model and various pertinent constraints, a constrained optimization problem is suitably formulated on the basis of the iLQR method. Next, we propose the implementation of the ADMM approach to split the optimization problem into several sub-problems, and these sub-problems can be efficiently solved. As a result, real-time computation and implementation can be realized through this framework, and thus it provides additional safety to the on-road driving tasks. Moreover, an illustrative example of autonomous driving is used to validate the performance of the approach in motion planning tasks, where different typical driving tasks have been scheduled as part of the test itinerary. As illustrated from the comparative computation times and simulation results, the significance as claimed in this work is suitably demonstrated.

			\bibliographystyle{IEEEtran}
			\bibliography{IEEEabrv,Reference}

		\end{document}